%% file: main.tex
\documentclass[10pt,twocolumn,letterpaper]{article}

\usepackage[pagenumbers]{iccv} % To force page numbers, e.g. for an arXiv version


\definecolor{iccvblue}{rgb}{0.21,0.49,0.74}
\usepackage[pagebackref,breaklinks,colorlinks,allcolors=iccvblue]{hyperref}
\usepackage{multirow}
\usepackage{graphicx}
\usepackage{changepage}  % 推荐这个包
\usepackage{colortbl} % For background color in table rows
\usepackage{xcolor}
\usepackage{makecell} % For allowing cell line breaks
\usepackage{array} 
\usepackage{pifont}
\usepackage{epigraph}

\title{UnitedVLN: Generalizable Gaussian Splatting for Continuous Vision-Language Navigation}

\author{%
Guangzhao Dai$^{1}$,
Jian Zhao$^{2}$\textsuperscript{*},
Yuantao Chen$^{3}$,
Yusen Qin$^{4}$,
Hao Zhao$^{4}$,
Guosen Xie$^{1}$, 
Yazhou Yao$^{1}$, \\
Xiangbo Shu$^{1}$\textsuperscript{*}, 
Xuelong Li$^{2}$\textsuperscript{*}\\
$^1$Nanjing University of Science and Technology \quad $^2$Northwest Polytechnical University\\
$^3$The Chinese University of Hong Kong, Shenzhen \quad $^4$Tsinghua University\\
}

\usepackage{epigraph}
\begin{document}
\maketitle

\epigraph{\textit{``The world as we have created it is a process of our thinking. It cannot be changed without changing our thinking.''}}{--- \textit{Albert Einstein}}

% 摘要自润-v5
\begin{abstract}
Vision-and-Language Navigation (VLN), where an agent follows instructions to reach a target destination, has recently seen significant advancements. In contrast to navigation in discrete environments with predefined trajectories, VLN in Continuous Environments (VLN-CE) presents greater challenges, as the agent is free to navigate any unobstructed location and is more vulnerable to visual occlusions or blind spots. Recent approaches have attempted to address this by imagining future environments, either through predicted future visual images or semantic features, rather than relying solely on current observations. However, these RGB-based and feature-based methods suffer from high-level semantic details or intuitive appearance-level information crucial for effective navigation. To overcome these limitations, we introduce a novel, generalizable 3DGS-based pre-training paradigm, called \textbf{UnitedVLN}, which enables agents to better explore future environments by unitedly predicting high-fidelity 360° visual images and semantic features. UnitedVLN employs two key schemes: search-then-query sampling and separate-then-united rendering, which facilitate efficient exploitation of neural primitives, helping to integrate both appearance and semantic information for more robust navigation. Extensive experiments demonstrate that UnitedVLN outperforms state-of-the-art methods on existing VLN-CE benchmarks.
\end{abstract}

\begin{figure}[t!]
    \centering
    \includegraphics[width=\linewidth]{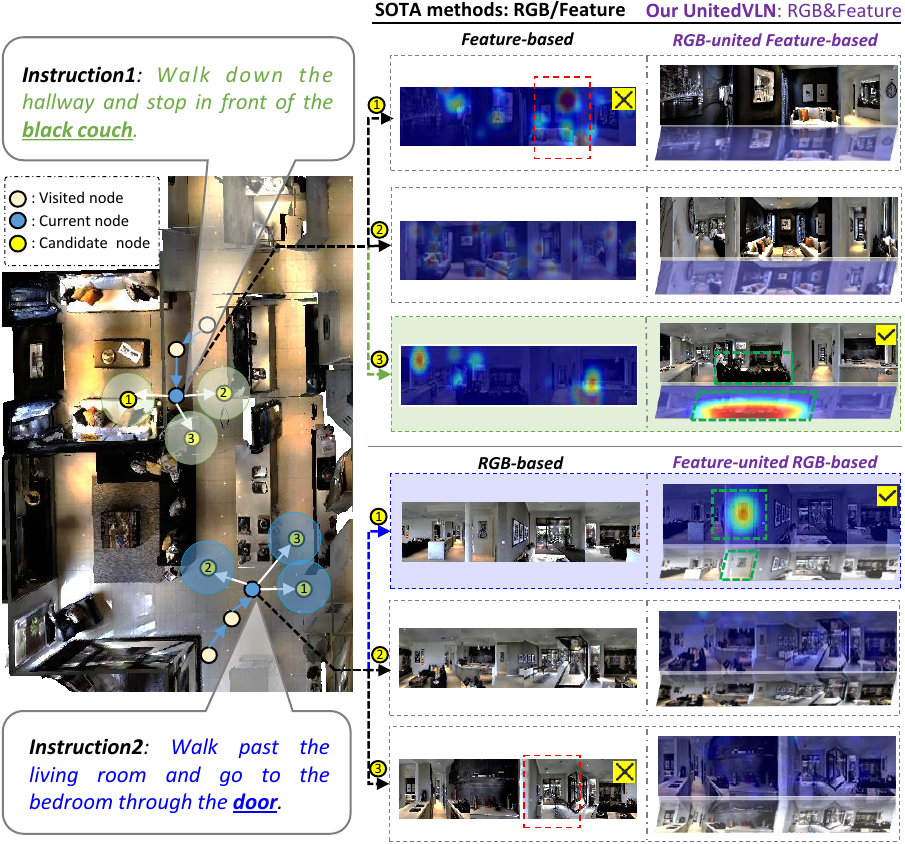}
    \caption{Main insights of UnitedVLN for VLN-CE: Unlike existing state-of-the-art methods that explore only either predicted images or features in future environments, our UnitedVLN fully integrates navigation cues from both appearance and semantic information. By leveraging two complementary rendering strategies—(1) appearance-level rendering (\textit{e.g.}, distinct colors) and (2) semantic-level rendering (\textit{e.g.}, appearance-similar elements like doors)—UnitedVLN enhances the agent's ability to interpret instructions accurately and navigate complex spaces. The visualizations show how UnitedVLN's combined approach results in more accurate route choices, reducing errors caused by occlusions or ambiguities that challenge purely RGB- or feature-based methods.}
    \label{fig1}
\end{figure}

\section{Introduction}
\label{sec:intro}

% 第一段，介绍通用的VLN任务。与横穿在具有完美预定义导航graph通用的VLN任务不同，执行VLN-CE要求agent在3D环境中用了low-actions进行自由导航。这个特性导致VLN-CE更加容易移动到视觉遮挡或盲区。
Vision-and-Language Navigation (VLN) \cite{Krantz2020r2r-ce, chen2022think-GL, chen2021history, VLN_2018vision} requires an agent to understand and follow natural language instructions to reach a target destination. This task has recently garnered significant attention in embodied AI \cite{2020reverie, 2020_RXR}. Unlike traditional VLN, where the agent navigates a predefined environment with fixed pathways, Continuous Environment VLN (VLN-CE) \cite{an2023etpnav, Wang2024lookahead} presents a more complex challenge. In VLN-CE, the agent is free to move to any unobstructed location, making low-level actions, such as moving forward 0.25 meters or turning 15 degrees. Consequently, the agent faces a higher risk of getting stuck or engaging in unproductive navigation behaviors, such as repeatedly hitting obstacles or oscillating in place, due to visual occlusions or blind spots in future environments. To address these challenges, recent methods \cite{jialu2023vlnsig, Wang2024lookahead,wang2023dreamwalker, koh2021pathdreamer} have focused on anticipating future environments, moving beyond reliance on current observations. In terms of synthesizing future observations at unvisited waypoints, these approaches, which include future visual images~\cite{wang2023dreamwalker, koh2021pathdreamer} and semantic features~\cite{jialu2023vlnsig, Wang2024lookahead}, can be categorized into two main explored paradigms: RGB-based and feature-based explorations. However, both explorations often fail to integrate intuitive appearance-level information with semantic context needed for robust navigation in complex environments. % Therefore, in this paper, we aim to equip VLN-CE model with both the capability to predict future visual images and also benefit from learning semantic details in future predicted observations. % 直接交代研究动机。之前的版本，采用”娓娓道来“，在引言倒数第二段才明确研究动机。

\textbf{RGB-based Exploration: exploring future visual images.} A straightforward approach to explore future environments is to predict images of future scenes, as images contain rich, appearance-level information (\textit{e.g.}, color, texture, and lighting) that is crucial for holistic scene understanding. Building on this idea, some methods~\cite{wang2023dreamwalker,koh2021pathdreamer} synthesize multiple future navigation trajectories by training an image generator~\cite{Isola2021gan} to predict panoramic images for path planning, demonstrating promising results. However, these methods are prone to suffering from the risk of lacking more complex, high-level semantic details. As illustrated in the bottom panel of Figure~\ref{fig1}, the agent struggles to distinguish the high-level semantic differences between objects like a ``door'' and a ``bedroom'', which may appear visually similar in the context of instructions. 
% This lack of semantic understanding often leads to poor decision-making.
 
\textbf{Feature-based Exploration: exploring future semantic features.} Rather than generating panoramic images, recent work~\cite{Wang2024lookahead} leverages a pre-trained Neural Radiance Field (NeRF)~\cite{mildenhall2021nerf} model to predict refined future semantic features through ray-to-pixel volume rendering. However, relying on rendered features alone can result in a lack of intuitive appearance information (\textit{e.g.}, color, texture, or lighting), which helps the agent to build stable knowledge about environments, such as general room layout rules and geometric structure~\cite{wang2023dreamwalker}. This lack potentially leads to more severe navigation errors. As shown at the top panel of Figure~{\color{red}\ref{fig1}}, the agent fails to accurately ground the color ``black" in the phrase ``black couch" when it encounters two differently colored couches, even though they are easily distinguishable at the visual appearance level. 

% This development motivates us to explore 3DGS to anticipate future scenarios during navigation, making agent behaviors more interpretable in each decision-making. Building on this, an effective rendering strategy that can simultaneously synthesize refined semantic details and holistic appearance information is also needed to be explored, for supporting robust future observations. 

In fact, human perception of an unknown environment is generally understood as a combination of appearance-level intuitive information and high-level semantic understanding, as suggested by studies in cognitive science~\cite{forrester1971counterintuitive,johnson1983mental,johnson2010mental}. Based on this insight, an agent designed to simulate human-like perception could also interpret instructions and navigate unseen environments. Recently, 3D Gaussian Splatting (3DGS)~\cite{kerbl20233d} has emerged in the computer graphics community for scene reconstruction, utilizing 3D Gaussians to speed up image rendering through a tile-based rasterizer. This motivates us to explore 3DGS to equip VLN-CE agent with both the capability to anticipate holistic future scenes while benefiting from learning semantic details in generated future observations. Furthermore, while NeRF is slower in image rendering than 3DGS, NeRF facilitates accurate semantic understanding, benefiting from volume rendering containing ray-to-pixel mechanism, which provides a precise one-to-one correspondence between sampled ray and pixel~\cite{liu2024mvsgauss}. Thus, it motivates us to explore an effective rendering strategy for integrating semantic understanding in volume rendering with Gaussian rendering.

% 虽然NeRF相较于3DGS在图像渲染速度较慢，但是受益于光线-像素的体渲染机制所带来的采样光线和像素一对一对应，NeRF可以促进精确细粒度语义理解。一种有效的渲染策略，可以高斯渲染与体渲染技术整合，也值得被探索。

Based on the above analysis, we propose a generalizable 3DGS-based paradigm, \textit{aka} \textbf{UnitedVLN}, which simultaneously renders both visual images and semantic features at higher quality (360° views) from sparse neural points, enabling the agent to more effectively explore future environments in VLN-CE. UnitedVLN primarily consists of two key components. First, we exploit a Search-Then-Query (STQ) sampling scheme for efficient neural point selection. For any neural points in the feature/point cloud, the scheme searches for neighboring points and queries their K-nearest neighbors. Second, to enhance navigation robustness, we introduce a Separate-Then-United (STU) rendering scheme. It utilizes hybrid rendering that integrates an efficient volume rendering with 3DGS to render high-level semantic features and visual images with appearance-level information. Our main contributions are highlighted as follows:
\begin{itemize}
\vspace{+2pt}
    \item \textbf{Unified VLN-CE Pre-training Paradigm}: We propose UnitedVLN, a generalizable 3DGS-based pre-training framework. It simultaneously renders both high-fidelity 360° visual images and semantic features from sparse neural primitives, enabling the agent to effectively explore future environments in VLN-CE.
    \par
    
    \item \textbf{Search-Then-Query Sampling Scheme}: We present a Search-Then-Query (STQ) sampling scheme for efficient selection of neural primitives. For each 3D position, the scheme searches for neighboring points and queries their K-nearest neighbors, improving model efficiency and resource utilization.
    \par
    
    \item \textbf{Separate-Then-United Rendering Scheme}: We present a Separate-Then-United (STU) rendering scheme that integrates an efficient volume rendering for predicting high-level semantic features in NeRF and visual images with appearance-level information in 3DGS, thereby enhancing the model's robustness in complex scenes.
    \par
\end{itemize}


\begin{figure*}[t!]
    \centering
    \includegraphics[width=\linewidth]{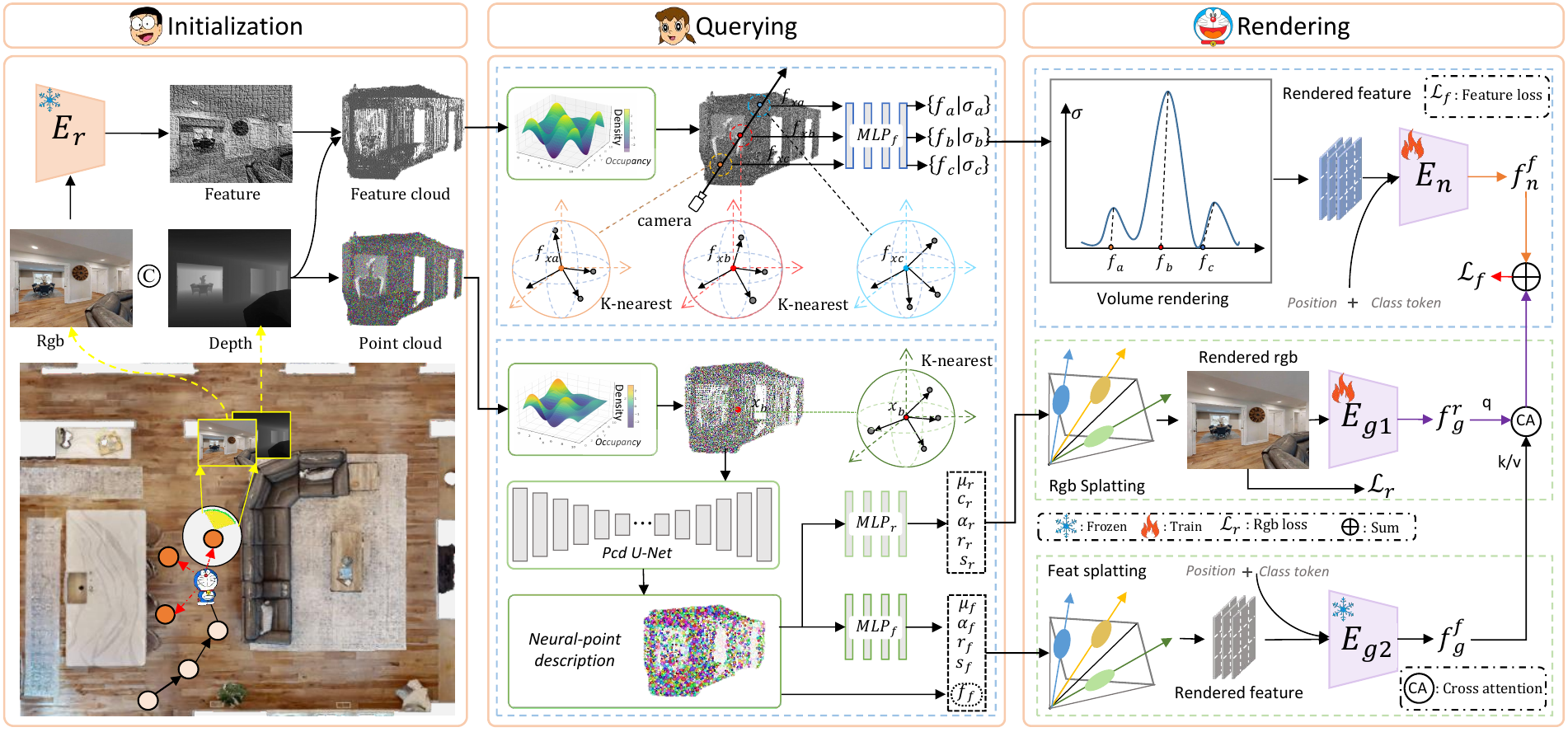}
    \caption{Overall framework of \textbf{UnitedVLN}. UnitedVLN obtains full higher-fidelity 360° visual observations, \textit{i.e.}, visual images and semantic features, through three stages: Initialization, Querying, and Rendering. In Initialization, it encodes the existing observed environments, \textit{i.e.}, visited and current observations, into the point cloud and feature cloud. In Querying, it adopts a Search-Then-Query sampling (STQ) scheme for efficient neural points sampling. Specifically, for any neural points in the feature/point cloud, it searches for each point in its neighborhood and queries its K-nearest points. Then, the sampled neural points in the feature/point cloud are fed into MLP to regress neural radiance, volume density, and images/feature {Gaussians}, respectively. In Rendering, for the neural radiance and images/feature {Gaussians} of the previous stage, it adopts a separate-then-united rendering (STU) scheme to render semantic features with high-level information via NeRF, and the visual image (interacted by 3DGS-rendered feature) with appearance-level information via 3DGS. Finally, the NeRF-rendered features and 3DGS-rendered image are integrated to ahead represent the semantic information in future environments.}
    \label{fig2}
    \vspace{-12pt}
\end{figure*}

\section{Related Work}
\label{related work}
\paragraph{Vision-and-Language Navigation (VLN).}
{VLN~\cite{VLN_2018vision,2020reverie,2020_RXR, Krantz2020r2r-ce} recently has achieved significant advance and increasingly introduced several proxy tasks, \textit{e.g.}, step-by-step instructions~\cite{VLN_2018vision,2020_RXR}, dialogs-based navigation~\cite{thomason2020cvdn}, and object-based navigation~\cite{2020reverie,zhu2021soon}, \textit{et al}. Among them, 
VLN in the Continuous Environment (VLN-CE)~\cite{VLN_2018vision,2018-speaker,tan2019learning,2019reinforced,hong2021vln-bert} entails an agent following instructions freely move to the target destination in a continuous environment. Similar to the general VLN in a discrete environment with perfect pathways, many previous methods of VLN-CE are focused on the visited environment~\cite{pashevich2021episodic,chen2021history,georgakis2022cm2,chen2022weakly,huang23vlmaps}, neglecting the exploration of the future environment, causing poor performance of navigation. Thus, some recent works~\cite{Wang2024lookahead,wang2023dreamwalker,koh2021pathdreamer} attempt to ahead explore the future environment instead of current observations, \textit{e.g.}, future RGB images or features. However, these RGB-based or feature-based methods rely on single-and-limited future observations~\cite{ramesh2021zero,li2023improving,wang2020active,feng2022uln}, lacking appearance-level intuitive information or high-level complicated semantic information.
Different from them, we propose a 3DGS-based paradigm named UnitedVLN that obtains full higher-fidelity 360° visual observations (both visual images and semantic features) for VLN-CE.
}
\vspace{-15pt}
\paragraph{3D Scene Reconstruction.}
{Recently, neural scene representations have been introduced \cite{mildenhall2021nerf,park2021nerfies,kerbl20233d}, such as Neural Radiance Fields (NeRF)~\cite{mildenhall2021nerf} and 3D Gaussian Splitting (3DGS)~\cite{kerbl20233d}, advancing the progress in 3D scene reconstruction. The more details can be found in~\cite{3d2022survey}. Thus, it attracts significant attention in the embodied AI community~\cite{devries2021unconstrained,kwon2023renderable} and extends their tasks with NeRF. However, NeRF typically requires frequent sampling points along the ray and multiple accesses to the global MLPs, which heavily intensifies the rendering time and makes it more challenging to generalization to unseen scenes. More recently, 3D Gaussian Splitting (3DGS)~\cite{kerbl20233d} utilizes 3D Gaussians with learnable parameters, speeding up image rendering through a tile-based rasterizer. It motivates us to leverage 3DGS to boost image rendering. To this end, we design a 3DGS-based VLN-CE model, which fully explores future environments and generalizes an agent to unite understanding of appearance-level intuitive information and high-level semantic information.}
\vspace{-10pt}
\section{Method}
\label{method}
\vspace{-2pt}
\paragraph{Task Setup.}
UnitedVLN focuses on the VLN-CE~\cite{Krantz2020r2r-ce,2020_RXR} task, where the agent is required to follow the natural instructions to reach the target location in the Continuous Environment. The action space in VLN-CE consists of a set of low-level actions, \textit{i.e.}, turn left 15 degrees, turn right 15 degrees, or move forward 0.25 meters. Following the standard panoramic VLN-CE setting~\cite{krantz2022sim2sim,Hong2022bridging,an2023etpnav}, 
at each time step $t$, the agent
receives a 360° panoramic observations that consists of 12 RGB images $\boldsymbol{\mathcal{I}}_{t}=\{\boldsymbol{r}_{t,i}\}_{i=1}^{12}$ and 12 depth images $\boldsymbol{\mathcal{D}}_{t}=\{\boldsymbol{d}_{t,i}\}_{i=1}^{12}$ surrounding its current location (\textit{i.e.}, 12 view images with 30° each separation). For each episode, it also receives a language instruction $\boldsymbol{\mathcal{W}}$, the agent needs to understand $\boldsymbol{\mathcal{W}}$, utilize panoramic observations of each step, and move to the target destination.
\paragraph{Overview of UnitedVLN.}
The framework of the proposed UnitedVLN is shown in Figure~\ref{fig2}. It is mainly through three stages, \textit{i.e.}, Initialization, Querying, and Rendering. In Initialization, it encodes the existing observed environments (\textit{i.e.}, visited and current observations) into the point cloud and feature cloud (§~\ref{sec:3.1}).  In Querying, it 
efficiently regresses feature radiance and volume density in NeRF and images/feature Gaussians in 3DGS, with the assistance of the proposed Search-Then-Query sampling (STQ) (§~\ref{sec_3.2}). In Rendering, it renders high-level semantic features via volume rendering in NeRF and appearance-level visual images via splitting in 3DGS, through separate-then-united rendering (STU) (§~\ref{sec_3.3}). Finally, the NeRF-rendered features and 3DGS-rendered features are aggregated to obtain future environment representation for navigation.

\vspace{-5PT}
\subsection{Neural Point Initialization}
\label{sec:3.1}
During navigation, the agent gradually stores visual observations of each step online, by projecting visual observations, including current nodes and visited nodes, into point cloud $\mathcal{B}$ and feature cloud $\mathcal{M}$. Among them, $\mathcal{M}$ is utilized to render high-level semantic features while $\mathcal{B}$ uses more dense points with color to render intuitive appearance-level information. Meanwhile, at each step, we also use a pre-trained waypoint predictor~\cite{Hong2022bridging} to predict navigable candidate nodes, following the practice of prior VLN-CE works~\cite{an2023bevbert,wang2023dreamwalker,an2023etpnav}. Note that $\mathcal{B}$ and $\mathcal{M}$ share a similar way of construction just different in projected subjects, \textit{i.e.}, images ($\mathcal{B}$) and feature map ($\mathcal{M}$). Here, we omit the construction of $\mathcal{M}$ for sake of readability. \textit{Please see for more details in supplementary material about the construction of }$\mathcal{M}$.

Point Cloud $\mathcal{B}$ stores holistic appearances of observed environments (\textit{e.g.}, current node and visited nodes), which consists of pixel-level point positions and colors, as shown in Figure~\ref{fig2}. 
% To encode the observed appearances of environments, we require taking each step image and depth for projection. 
Specifically, at each time step $t$, we first use 12 RGB images $\boldsymbol{\mathcal{I}}_{t}=\{\boldsymbol{r}_{t,i}\}_{i=1}^{12}$ to enumeratively project pixel colors $\{\boldsymbol{c}_{t,i}\in\mathcal{\mathbb{R}}^{H\times W\times 3}\}_{i=1}^{12}$, where $H\times W\times 3$ denotes RGB images resolution. 
For the sake of calculability, we omit all the subscripts $(i,h,w)$ and denoted it as $l$, where $l$ ranges from 1 to $L$, and $L=12 \cdot$$H$$\cdot$$W$. Then, we use $\boldsymbol{\mathcal{D}}_{t}=\{\boldsymbol{d}_{t,i}\}_{i=1}^{12}$ to obtain the point positions. Through camera extrinsic $[\mathbf{R}, \mathbf{T}]$ and intrinsics $\mathbf{K}$,
each pixel in the $i$-view image $\boldsymbol{r}_{t,i}$ is mapped to its 3D world position $\boldsymbol{P}_{t,l}=[p_x,p_y,p_z]$ using depth images $\boldsymbol{d}_{t,i}$, as 
\vspace{-5pt}
\begin{equation}
    \begin{split}
    \boldsymbol{P}_{t,l}
    &= 
    \begin{bmatrix}\boldsymbol{d}_{t,l} \mathbf{R}^{-1} \mathbf{K}^{-1} 
    \begin{bmatrix}
    h \ w \ 1
    \end{bmatrix}^T
    - \mathbf{T} \;
    \end{bmatrix}^T.
    \end{split}\label{maping:projection}
\end{equation} 

\vspace{-4pt}
Based on this, in each step, we gradually perceive point colors and their positions into the point cloud $\mathcal{B}$, as
\vspace{-2pt}
\begin{equation}
    \label{eq:B}
    \mathcal{B}_{t}= \mathcal{B}_{t-1} \cup \{[\boldsymbol{P}_{t,j}, \boldsymbol{C}_{t,j}]\}_{j=1}^{J}.
    \end{equation}

Similar to Eq.~\ref{maping:projection}, we project feature maps extracted by pre-trained CLIP-ViT-B~\cite{radford2021learning} to obtain feature cloud $\mathcal{M}$, as
\vspace{-5pt}
\begin{equation}
    \label{eq:fcd}
    \mathcal{M}_{t}= \mathcal{M}_{t-1} \cup \{[\boldsymbol{Q}_{t,u},\boldsymbol{H}_{t,u}, \boldsymbol{\theta}_{t,u},\boldsymbol{s}_{t,u}]\}_{j=1}^{U}.
    \end{equation}
Here, $\boldsymbol{Q}_{t,u}$ and $\boldsymbol{H}_{t,u}$ denote 3D positions and grid embedding in feature map, respectively. The $\boldsymbol{\theta}_{t,u}$ and $\boldsymbol{s}_{t,u}$ denote feature point directions and feature scales, respectively.
\vspace{-3pt}
\subsection{Search-Then-Query Sampling}
\label{sec_3.2}
\vspace{-3pt}
\paragraph{Sampling in 3DGS.}
To obtain more representative points, we filter low-information or noisy points in the source point cloud $\mathcal{B}$ by two steps, \textit{i.e.}, point search and point query. In the point search, we construct a KD-Tree~\cite{grandits_geasi_2021} to represent point occupancy via coarse grids. Initialize with threshold $\epsilon$ to detect K nearest points by the matrix of distance $\boldsymbol{D}_{\text{kn}}$:
\vspace{-3pt}
\begin{equation}
    \label{eq: k-neast}
    {\boldsymbol{D}_{\text{kn}}} = \text{D}_\text{oc}^{\text{tree}} \left( \left\{ {p}_i \in {P} \;\middle|\; d_{\text{oc}}({p}_i) < \epsilon^2 \right\}, \text{K} \right),
    \vspace{-3pt}
\end{equation}
where $\boldsymbol{d}_{\text{oc}}(\boldsymbol{p}_i)$ denotes the square distance of $\boldsymbol{p}_i$ the nearest neighbor grid point, and $\text{D}_\text{oc}^{\text{tree}}$ denotes K-nearest search from occupancy tree. For these satisfied points filtered by Eq.~\ref{eq: k-neast}, we calculate the total distance (\textit{i.e.}, density) to its neighbored points. Then, we query points with local maxima in the density distribution, which represent the most representative and dense structural information. In this way, the dense points $\boldsymbol{P}$ is reformulated to sparse new points $\boldsymbol{P}^{'}$:
\begin{equation}
    \label{eq: sample points}
    \boldsymbol{P}^{'} = \left\{ \boldsymbol{q}_i \;\middle|\; i \in \Gamma \left(\frac{1}{\sum_{j} \boldsymbol{D_{ij}}}\right) 
    \cup \Lambda \left(\frac{1}{\sum_{j} \boldsymbol{D_{ij}}}\right) \right\},
\end{equation}
where $\boldsymbol{D}_{ij}$ denotes the distance between the $i$-th query point and its $j$-th neighbor. The $\Gamma$ and $\Lambda$ denote density and peak selection functions, respectively.

% Thus, with sampled points $P^{'}$ and their corresponding colors $\boldsymbol{C}^{'}$, we obtain more representative point cloud $\mathcal{B}^{'}= \{ P^{'}, C^{'} \} $, which are then to regress image/feature Gaussians via 3DGS. 

Based on $P^{'}$ and their colors $\boldsymbol{C}^{'}$, we use a multi-input and single-output UNet-like architecture~\cite{wang2024pfgs} to encode points with different scales to obtain neural descriptors $\boldsymbol{F}_p^{'}$. Then, we regress it to several Gaussian properties: rotation $\boldsymbol{R} \in \mathrm{R}^{4}$, scale factor $\boldsymbol{S} \in \mathrm{R}^{3}$ and opacity $\alpha \in [0,1]$, as
\vspace{-3pt}
\begin{adjustwidth}{0pt}{0pt}
    \setlength{\abovedisplayskip}{-3pt}%
    \setlength{\abovedisplayshortskip}{\abovedisplayskip}%
    \begin{equation}
    \boldsymbol{R},\boldsymbol{S}, \alpha = \texttt{N}(\mathcal{H}_\text{r}(\boldsymbol{F}_{p}^{'})),\texttt{E}(\mathcal{H}_\text{s}(\boldsymbol{F}_p^{'})),\texttt{S}(\mathcal{H}_\text{o}(\boldsymbol{F}_p^{'})).
    \label{eq: regressor}
    \end{equation}
    \vspace{-16pt}
\end{adjustwidth}
Here $\mathcal{H}_\text{r}$, $\mathcal{H}_\text{s}$, and $\mathcal{H}_\text{o}$ denote three different MLPs to predict corresponding properties of Gaussian. The $\texttt{N}$, $\texttt{E}$, and  $\texttt{S}$ denote normalization operation, exponential function, and sigmoid function, respectively. {We also use the neural descriptors $\boldsymbol{P}^{'}$ to replace colors to obtain feature Gaussians, following point-rendering practice~\cite{wang2024pfgs}.} Thus, the images Gaussians $\mathcal{G_{\scalebox{0.6}{$\triangle$}}}$ and feature Gaussians $\mathcal{G_{\scalebox{0.8}{$\triangledown$}}}$ are formulated as
\vspace{-3pt}
\begin{equation}
    \mathcal{G_{\scalebox{0.6}{$\triangle$}}},\mathcal{G_{\scalebox{0.8}{$\triangledown$}}} = \{\boldsymbol{R}, \boldsymbol{S}, \alpha, \boldsymbol{P}^{'}, \boldsymbol{C}{'}  \},\{\boldsymbol{R}, \boldsymbol{S}, \alpha, \boldsymbol{P}^{'}, \boldsymbol{F}_{p}^{'} \}.
    \label{eq: regressor}
\end{equation}

\vspace{-16pt}
\paragraph{Sampling in NeRF.} For each point $q$ along the ray, we then use KD-Tree~\cite{grandits_geasi_2021} to search k-nearest features $\{\boldsymbol{h}_{k}\}_{k=1}^K$ within a certain radius $R$ in $\mathcal{M}$. Based on $\{\boldsymbol{h}_{k}\}_{k=1}^K$, for the feature point $\boldsymbol{p}_q$, we use a MLP $\varphi_\text{1}$ to aggregate a new feature vector $\boldsymbol{f}_{q}^{'}$ that represent point $\boldsymbol{q}$ local content as,
\begin{adjustwidth}{0pt}{0pt}
    \setlength{\abovedisplayskip}{-5pt}%
    \setlength{\abovedisplayshortskip}{\abovedisplayskip}%
    \begin{equation}
    \boldsymbol{f}_{q,h} = \varphi_\text{1}(\boldsymbol{f}_{q}, \boldsymbol{p}_q - \boldsymbol{h}_{k}),
    \end{equation}
\end{adjustwidth}
\vspace{-5pt}
\begin{adjustwidth}{0pt}{0pt}
    \setlength{\abovedisplayskip}{-7pt}%
    \setlength{\abovedisplayshortskip}{\abovedisplayskip}%
    \begin{equation}
     \boldsymbol{f}_{q}^{'} = \sum_{k}^{K}  s_{k} \frac{w_{k}}{\sum w_{k}}  \boldsymbol{f}_{q,h}, \ w_{k} = \frac{1}{\| \boldsymbol{h}_{k} -  \boldsymbol{p}_q\|}.
    \label{randiance}
    \end{equation}
\end{adjustwidth}
Here, $w_k$ denotes inverse distance, which makes closer neural points contribute more to the sampled point computation, $s_k$ denotes feature scales (\textit{cf.} in Eq.~\ref{eq:fcd}),
$\boldsymbol{f}_q$ denotes feature embedding of $ \boldsymbol{p}_q$, and $\boldsymbol{h}_{k} -  \boldsymbol{p}_q$ denotes the relative position of $\boldsymbol{p}_q$ to $\boldsymbol{h}_{k}$. Then, through two MLPs $\varphi_\text{2}$ and $\varphi_\text{3}$, we regress the view-dependent feature radiance $r$ and $\sigma$ with the given view direction 
$\theta_{k}$ and feature scale $s_k$ of $\boldsymbol{h}_{k}$.
\vspace{-6pt}
\begin{equation}
r = \varphi_\text{2}(\boldsymbol{f}_{q}^{'}, \theta_{k}),
\label{eq:radiance}
\end{equation}
\vspace{-13pt}    
\begin{adjustwidth}{0pt}{0pt}
    \centering
    \setlength{\abovedisplayskip}{-5pt}%
    \setlength{\abovedisplayshortskip}{\abovedisplayskip}%
    \begin{align}
    \sigma &= \sum_{i} \varphi_\text{3}\left(\boldsymbol{f}_{q,h}\right) \gamma_i \frac{w_i}{\sum_{i} w_i}, \quad w_i = \frac{1}{\left\| \boldsymbol{h}_k - \boldsymbol{p}_q \right\|}. \label{eqn:weighted_density}
    \end{align}
\end{adjustwidth}

\subsection{Separate-Then-United Rendering}
\label{sec_3.3}
\vspace{-2pt}
As shown in Figure~\ref{fig2}, we render future observations, \textit{i.e.}, the feature map (in NeRF branch), image and feature map (in 3DGS branch). Specifically, by leveraging the differentiable rasterizer, we first render image/feature Gaussian parameters (\textit{cf.} in Eq.~\ref{eq: regressor}) to image/feature map $\boldsymbol{I}_{g}$/$\boldsymbol{F}_{g}$ as,
\vspace{-3pt}
\begin{equation}
    \boldsymbol{I}_{g},\boldsymbol{F}_{g} = Raster \{ \mathcal{G_{\scalebox{0.6}{$\triangle$}}} \},Raster \{ \mathcal{G_{\scalebox{0.8}{$\triangledown$}}} \}.
    \label{eq: render-image}
    \vspace{-3pt}
\end{equation}

Note that $\boldsymbol{I}_{g}$ can maintain well quality with sufficient points. However, when the point sparsely distributes over the space, the representation ability of corresponding Gaussians is constrained, especially in complicated local areas, resulting in low-quality visual representation~\cite{wang2024pfgs}. In contrast, $\boldsymbol{F}_{g}$ contains rich context information to describe local geometric structures with various scales. In other words, $\boldsymbol{F}_{g}$ can be treated as a complement to $\boldsymbol{I}_{g}$. Therefore, we use multi-head cross-attention (CA) to replenish context from $\boldsymbol{F}_{g}$ into $\boldsymbol{I}_{g}$, for promoting representation. To sum up, the aggregation representation $\boldsymbol{f}_{g}^{rf}$ can be formulated as,
\vspace{-3pt}
\begin{equation}
\label{gs-features}
    \boldsymbol{f}_{g}^{rf} = \text{CA}(\varphi_\text{g1}(\boldsymbol{I}_{g}),\varphi_\text{g2}(\boldsymbol{F}_{g}),\varphi_\text{g2}(\boldsymbol{F}_{g})),
    \vspace{-3pt}
\end{equation}
where $\varphi_\text{g1}$ and $\varphi_\text{g2}$ denote two visual encoder of CLIP-ViT-B~\cite{radford2021learning} for encoding image and feature map.

In NeRF branch, with the obtained feature radiance $r$ and volume density $\sigma$ of sampled points along the ray, we render the future feature $\boldsymbol{F}_{f}$ by using the volume rendering~\cite{mildenhall2021nerf}. Similarly, we use the encoder $\varphi_\text{f}$ of Transformer~\cite{vanswani2017transformer} to extract the feature embedding $\boldsymbol{f}_n$ for $\boldsymbol{F}_{f}$ in NeRF, as
\vspace{-5pt}
\begin{equation}
\label{nerf-features}
    \boldsymbol{f}_{n}^{f} = \varphi_\text{f}(\boldsymbol{F}_f).
\end{equation}

\vspace{-5pt}
To improve navigation robustness, we aggregate future feature embedding $\boldsymbol{f}_{g}^{rf}$ (\textit{cf.} in Eq.~\ref{gs-features}) and $\boldsymbol{f}_{n}^{f}$ (\textit{cf.} in Eq.~\ref{nerf-features}) as a view representation of future environment. In this way, we aggregate all 12 future-view embeddings $\boldsymbol{F}^{\text{future}}$ via average pooling and project them to a future node. Finally, we use a feed-forward network (FFN) to predict navigation scores between the candidate node $\boldsymbol{F}^{\text{candidate}}$ and the future node $\boldsymbol{F}^{\text{future}}$ in the topological map, following practices of previous methods~\cite{an2023etpnav,Wang2024lookahead}. Note that the scores for visited nodes are masked to avoid agent unnecessary repeated visits. Based on navigation goal scores, we select a navigation path with a maximum score, as 
\vspace{-2pt}
\begin{equation}
\label{eq: navigation-logits}
    S^{\text{path}} = \{\texttt{Max} ([\texttt{FFN}(\boldsymbol{F}^{\text{candidate}}),\texttt{FFN}(\boldsymbol{F}^{\text{future}})]\}.
\end{equation}

\subsection{Objective Function}
According to the stage of VLN-CE, UnitedVLN mainly has two objectives, \textit{i.e.}, one aims to achieve better render quality of images and features (\textit{cf.} Eq~\ref{eq: render-image}) in the pre-training stage and the other is for better navigation performance (\textit{cf.} Eq~\ref{eq: navigation-logits}) in the training stage. 
\textit{Please see in supplementary material for details about the setting of objective loss.}

% On the pre-training stage, we totally use four loss to constrain rendered images/features with ground-truth images/features: L1 loss $\mathcal{L}_{1}^{r}$, L2 loss $\mathcal{L}_{2}^{r}$, and SSIM loss $\mathcal{L}_{ssim}^{r}$ for constraining colors and geometric structures between rendered images and ground-truth images; L2 loss $\mathcal{L}_{2}^{f}$ for optimizing rendered feature with ground-truth features. On the training stage, we use cross-entropy loss $\mathcal{L}_{ce}$ to constrain the goal scores in Eq.~\ref{eq: navigation-logits}. 
% \vspace{-10pt}

\vspace{-3pt}
\section{Experiment}
\vspace{-3pt}
\label{experiment}

\begin{table*}

\tabcolsep=0.215cm
\centering{}%
\begin{tabular}{l||cccc|cccc|cccc}
\Xhline{2.5\arrayrulewidth} % 顶部加粗横线
\multirow{2}{*}{Methods} & \multicolumn{4}{c|}{Val Seen} & \multicolumn{4}{c|}{Val Unseen} & \multicolumn{4}{c}{Test Unseen}\tabularnewline
\cline{2-13} \cline{3-13} \cline{4-13} \cline{5-13} \cline{6-13} \cline{7-13} \cline{8-13} \cline{9-13} \cline{10-13} \cline{11-13} \cline{12-13} \cline{13-13}
 & NE\textdownarrow{} & OSR\textuparrow{} & SR\textuparrow{} & SPL\textuparrow{} & NE\textdownarrow{} & OSR\textuparrow{} & SR\textuparrow{} & SPL\textuparrow{} & NE\textdownarrow{} & OSR\textuparrow{} & SR\textuparrow{} & SPL\textuparrow{}\tabularnewline
\hline \hline 

CM$^{2}$\ \cite{georgakis2022cm2} & 6.10 & 51 & 43 & 35 & 7.02 & 42 & 34 & 28  & 7.70 & 39 & 31 & 24\tabularnewline

WS-MGMap\ \cite{chen2022weakly} & 5.65 & 52 & 47 & 43 & 6.28 & 48 & 39 & 34 & 7.11 & 45 & 35 & 28\tabularnewline

Sim-2-Sim\ \cite{krantz2022sim2sim} & 4.67 & 61 & 52 & 44 & 6.07 & 52 & 43 & 36 & 6.17 & 52 & 44 & 37\tabularnewline

ERG\textcolor{black}\ \cite{Ting2023graph-vlnce} & 5.04 & 61 & 46 & 42 & 6.20 & 48 & 39 & 35 & - & - & - & -\tabularnewline

CWP-CMA\textcolor{black} \ \cite{Hong2022bridging} & 5.20 & 61 & 51 & 45 & 6.20 & 52 & 41 & 36 & 6.30 & 49 & 38 & 33\tabularnewline

CWP-RecBERT\textcolor{black}\ \cite{Hong2022bridging} & 5.02 & 59 & 50 & 44 & 5.74 & 53 & 44 & 39 & 5.89 & 51 & 42 & 36\tabularnewline 

GridMM\textcolor{black}\
\cite{wang2023gridmm}& 4.21 & 69 & 59 & 51 & 5.11 & 61 & 49 & 41 & 5.64 & 56 & 46 & 39
\tabularnewline
 
Reborn\textcolor{black}\ \cite{an20221st} &  4.34 & 67 & 59 & 56 & 5.40 & 57 & 50 & 46 & 5.55 & 57 & 49 & 45\tabularnewline

Ego$^{2}$-Map\textcolor{black}\ \cite{hong2023learning} & - & - & - & - & 4.94 & - & 52 & 46 & 5.54 & 56 & 47 & 41 \tabularnewline

Dreamwalker\textcolor{black}\ \cite{wang2023dreamwalker} & 4.09 & 66 & 59 & 48 & 5.53 & 59 & 49 & 44 & 5.48 & 57 & 49 & 44 \tabularnewline

Energy\textcolor{black}\ \cite{liurui2024energy} & 3.90 & 73 & 68 & 59 & 4.69 & 65 & 58 & 50 & 5.08 & 64 & 56 & 48 \tabularnewline

BEVBert\textcolor{black}\ \cite{an2023bevbert} & - & - & - & - & 4.57 & 67 & 59 & 50 & 4.70 & 67 & \textbf{59} & \textbf{50}\tabularnewline

ETPNav\textcolor{black}\ \cite{an2023etpnav} &  3.95 & 72 & 66 & 59 & 4.71 & 65 & 57 & 49 & 5.12 & 63 & 55 & 48\tabularnewline

\textcolor{black} HNR {\cite{Wang2024lookahead}} & 3.67 & 76 & 69 & 61 & 4.42 & 67 & 61 & \textbf{51} & 4.81 & 67 & 58 & \textbf{50}\tabularnewline

\hline

\textcolor{black} UnitedVLN (Ours) & \textbf{3.30} & \textbf{78} & \textbf{70} & \textbf{62} & \textbf{4.29} & \textbf{70} & \textbf{62} & {\textbf{51}} & \textbf{4.69} & \textbf{68} & \textbf{59} & 49
 \tabularnewline
\hline 
\end{tabular}
\vspace{-5pt}
\caption{Evaluation on the R2R-CE dataset.}\label{R2R-CE_sota}
\vspace{-8pt}
\end{table*}

\begin{table*}
% \small
\tabcolsep=0.245cm
\centering{}%
\begin{tabular}{l||ccccc|ccccc}
\Xhline{2.5\arrayrulewidth} % 顶部加粗横线
 \multirow{2}{*}{Methods} &\multicolumn{5}{c|}{Val Seen} & \multicolumn{5}{c}{Val Unseen}\tabularnewline
\cline{2-11}
 & NE\textdownarrow{} & SR\textuparrow{} & SPL\textuparrow{} & NDTW\textuparrow{} & SDTW\textuparrow{} & NE\textdownarrow{} & SR\textuparrow{} & SPL\textuparrow{} & NDTW\textuparrow{} & SDTW\textuparrow{} \tabularnewline
\hline \hline 

CWP-CMA\textcolor{black}~\cite{Hong2022bridging} & - & - & - & - & - &  8.76 & 26.6 & 22.2 & 47.0 & - \tabularnewline

CWP-RecBERT\textcolor{black}~\cite{Hong2022bridging} & - & - & - & - & - & 8.98 & 27.1 & 22.7 & 46.7 & - \tabularnewline

Reborn\textcolor{black}~\cite{an20221st} & 5.69 & 52.4 & 45.5 & 66.3 & 44.5 & 5.98 & 48.6 & 42.1 & 63.4 & 41.8\tabularnewline

ETPNav\textcolor{black}~\cite{an2023etpnav} & 5.03 & 61.5 & 50.8 & 66.4 & 51.3 & 5.64 & 54.8 & 44.9 & 61.9 & 45.3\tabularnewline

HNR~\cite{Wang2024lookahead} & {4.85} & {63.7} & 53.2 & {68.8} & {52.8} & {5.51} & {56.4} & {46.7} & {63.6} & {47.2}
 \tabularnewline
\hline

\textcolor{black} UnitedVLN (Ours) & \textbf{4.71} & \textbf{64.9} & {\textbf{53.8}} & \textbf{69.9} & \textbf{53.5} & \textbf{5.49} & \textbf{57.7} & {\textbf{47.1}} & \textbf{64.0} & \textbf{47.8} 
 \tabularnewline
\hline 

\end{tabular}
\vspace{-5pt}
\caption{Evaluation on the RxR-CE dataset.}\label{RxR-CE_sota}
\end{table*}

\subsection{Datasets and Evaluation Metrics}
\paragraph{Datasets.} To improve the rendered quality of images and features, we first pre-train the proposed 3DGS-based UnitedVLN on the large-scale indoor HM-3D dataset. Following the practice of prior VLN-CE works~\cite{Hong2022bridging,wang2023dreamwalker,an2023etpnav}, 
we evaluate our UnitedVLN two VLN-CE public benchmarks, \textit{i.e,} R2R-CE~\cite{Krantz2020r2r-ce} and RxR-CE~\cite{2020_RXR}. 
\textit{Please see supplementary material for more details about the illustration of datasets.}
\vspace{-10pt}
\paragraph{Evaluation Metrics.} Following standard protocols in previous methods~\cite{Wang2024lookahead,an2023etpnav,wang2023dreamwalker}, we use several metrics in VLN-CE for evaluating our UnitedVLN performance of navigation: Navigation Error (NE), Success Rate (SR), Oracle stop policy (OSR), Normalized inverse of the Path Length (SPL), Normalized Dynamic Time Warping (nDTW), and Success weighted by normalized Dynamic Time Warping (SDTW).

\begin{table}[t!]
\vspace{-14pt}
\small
\tabcolsep=0.08cm
\centering{}%
\begin{tabular}{l||ccccc}
\Xhline{2.5\arrayrulewidth} % 顶部加粗横线
Methods  & NE\textdownarrow{} & OSR\textuparrow{} & SR\textuparrow{} & SPL\textuparrow{}\tabularnewline
\hline \hline 

A1 (Base)   & 4.73 & 64.9 & 57.6 & 46.5\tabularnewline

A2 (Base + STQ + NeRF Rendering) & 4.45 & 67.8 & 61.4 & 49.9 \tabularnewline

A3 (Base + STQ + 3DGS Rendering)  & 4.37 & 68.4 & 61.9 & 50.6 \tabularnewline
\hline
A4 (Base + STU)  & {4.31} & {69.3} & {62.0} & {50.4}\tabularnewline
A5 (Base + STQ + STU)  & \textbf{4.29} & \textbf{70.0} &\textbf{62.4} &\textbf{51.1}\tabularnewline

\hline 
\end{tabular}
\vspace{-5pt}
\caption{Ablation study of each component of UnitedVLN model.} \label{table:each-commpont}
\vspace{-10pt}
\end{table}
\vspace{-5pt}

\subsection{Implementation Details}
Our UnitedVLN adopts a pertaining-then-finetuning train paradigm, which first pre-training a generalized 3DGS to regress future observations, then generalizes it to the agent for VLN-CE in an inference way. \textit{Please see in supplementary material about settings of pre-training and training.}
\vspace{-3pt}
\subsection{Comparison to State-of-the-Art Methods}
Table~\ref{R2R-CE_sota} and \ref{RxR-CE_sota} %display 
show the performance of UnitedVLN compared with the state-of-the-art methods on the R2R-CE and RxR-CE benchmarks respectively. Overall, UnitedVLN achieves SOTA results in the majority of metrics, proving its effectiveness from diverse perspectives. As demonstrated in Table~\ref{R2R-CE_sota}, on the R2R-CE dataset, our method outperforms the SOTA method (\textit{i.e.}, HNR~\cite{Wang2024lookahead}): +1\% on SR and +3\% on OSR for the val unseen split; +1\% on SR and -1.2\% on NE for the test unseen split. Meanwhile, as illustrated in Table~\ref{RxR-CE_sota}, the proposed method also achieves improvements in the metrics on the RxR-CE dataset.

Specifically, compared with Dreamwalker~\cite{wang2023dreamwalker} that shares a partial idea of UnitedVLN to predict future visual images in Table~\ref{R2R-CE_sota}, our UnitedVLN model achieves performance gains of about 11\% on SR for all splits. Our UnitedVLN supplements the future environment with high-level semantic information, which is better than Dreamwalker depending on a single visual image. Compared with HNR, we still outperform its SR and OSR on the val unseen set, which relies on future features of environments but lacks appearance-level intuitive information. It proves that UnitedVLN can effectively improve navigation performance by uniting future appearance-level information (visual images) and high-level semantics (features).
\begin{figure*}[t!]
    \centering
    \includegraphics[width=\linewidth]{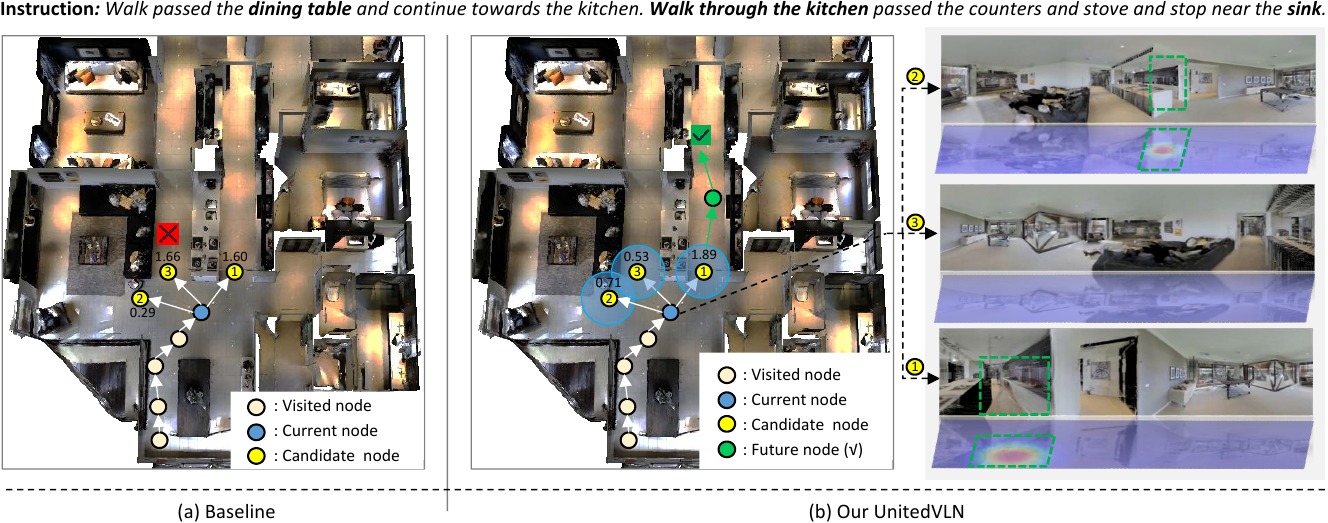}
    \caption{Visualization example of navigation strategy on the val unseen split of the R2R-CE dataset. (a) denotes the navigation strategy of the baseline model. (b) denotes the RGB-united-Feature exploration strategy of our unitedVLN.}
    \label{fig:vis}
\vspace{-10pt}
\end{figure*}

\vspace{-2pt}
\subsection{Ablation Study}
\vspace{-2pt}
{We conduct extensive ablation experiments to validate key designs of UnitedVLN. Results are reported on the R2R-CE val unseen split with a more complex environment and difficulty, and the best results are highlighted.} 
\vspace{-6pt}

\begin{table}[t!]
\vspace{-14pt}
\tabcolsep=0.19cm
\centering{}%
\begin{tabular}{l||ccccc}
\Xhline{2.5\arrayrulewidth} % 顶部加粗横线
Methods  & NE\textdownarrow{} & OSR\textuparrow{} & SR\textuparrow{} & SPL\textuparrow{}\tabularnewline
\hline \hline 

B1 (ETPNav)  & 4.71 & 64.8 & 57.2 & 49.2\tabularnewline

B2 (UnitedVLN$_{\text{ETPNav}}$)  & \textbf{4.25} & \textbf{67.6} & \textbf{58.9} & \textbf{49.8}\tabularnewline
\hline
B3 (HNR)  & 4.42 & 67.4 & 60.7 & \textbf{51.3}\tabularnewline

B4 (UnitedVLN$_{\text{HNR}}$)  & \textbf{4.29} & \textbf{70.1} & \textbf{62.4} & 51.1\tabularnewline

\hline 
\end{tabular}
\vspace{-5pt}
\caption{Generalization analysis of UnitedVLN among different VLN-CE models.} \label{table: generalization}
\vspace{-18pt}
\end{table}

\paragraph{Effect on each component of UnitedVLN.}
\vspace{-6pt}
In this part, we analyze the effect of each component in UnitedVLN. As illustrated in Table~\ref{table:each-commpont}, A1 (Base) denotes performance of baseline model.  Compared with A1, the performance gain of A2 is improved when equipped with future environmental features via NeRF rendering. Compared with A1, the navigation performance gain of A3 is also extended when equipped with 3DGS rendering on the Eq.~\ref{gs-features}, which validates the effectiveness of learning appearance-level information (visual images). Compared with A5, the performance of A4 (Base + STU) is decreased when remove point sampling in STQ. It proves that STQ can effectively enhance the performance of navigation by performing efficient neural point sampling. From the results in A5, when we combine features from NeRF rendering and 3DGS rendering on Section~\ref{sec_3.3} and efficient sampling in STQ, it further improves and achieves the best performance, by +5.1(\%) on OSR. To sum up, all components in UnitedVLN can jointly improve the VLN-CE performance.

\vspace{-5pt}
\begin{table}[t!]
\small
\tabcolsep=0.26cm
\centering
\begin{tabular}{l||cc}
\Xhline{2.5\arrayrulewidth} % 顶部加粗横线
Methods & Pre-train stage & Train stage \tabularnewline
\hline \hline
C1 (HNR)   & 2.21s (0.452Hz)   & 2.11s (0.473Hz) \tabularnewline    
C2 (UnitedVLN)   & \textbf{0.036s} (\textbf{27.78Hz})   & \textbf{0.032s} (\textbf{31.25Hz})     \tabularnewline
\hline 
\end{tabular}
\vspace{-7pt}
\caption{Runtime analysis measured on one NVIDIA RTX 3090.}
\label{tab: time}
\vspace{-8pt}
\end{table}

\begin{table}[t!]
\vspace{-5pt}
\small
\noindent\begin{minipage}[t]{1\columnwidth}%
\tabcolsep=0.18cm
\begin{center}
\begin{tabular}{l||ccccc}
\Xhline{2.5\arrayrulewidth} % 顶部加粗横线
 Extractors & 
NE\textdownarrow{} & OSR\textuparrow{} & SR\textuparrow{} & SPL\textuparrow{}\tabularnewline
\hline \hline

D1 (ViT-B/16-ImageNet~\cite{deng2009imagenet}) &  4.37 & 69.8 & 61.5 & 50.6 \tabularnewline
D2 (ViT-B/16-CLIP~\cite{radford2021learning}) &  \textbf{4.29} & \textbf{70.0} & \textbf{62.4} & \textbf{51.1} \tabularnewline
\hline 
\end{tabular}
\par\end{center}%
\end{minipage}
\vspace{-8pt}
\caption{Ablation study of the different extractors of 3DGS branch.}\label{tab: feature extractors}
\vspace{-15pt}
\end{table}

\vspace{-10pt}
\paragraph{Effect on numbers of K-nearest on point sampling.}
Figure~\ref{fig: K_neast} shows the effect of point sampling with different numbers of k-nearest features on SR accuracy. We set K $\in \{1, 8, 16, 32\}$ for investigation, and K stabilizes from 8 and converges to the best performance at 16. Here, we select K = 16 in NeRF/3DGS sampling. It 
can be found that when K is set to smaller than 16 or larger than 16, the accuracy of navigation decreases slightly. Nevertheless, a larger or smaller number in a moderate range is acceptable since contextual information is aggregated by K-nearest sampling.
\begin{figure*}[t!]
    \centering
    \includegraphics[width=.96\linewidth]{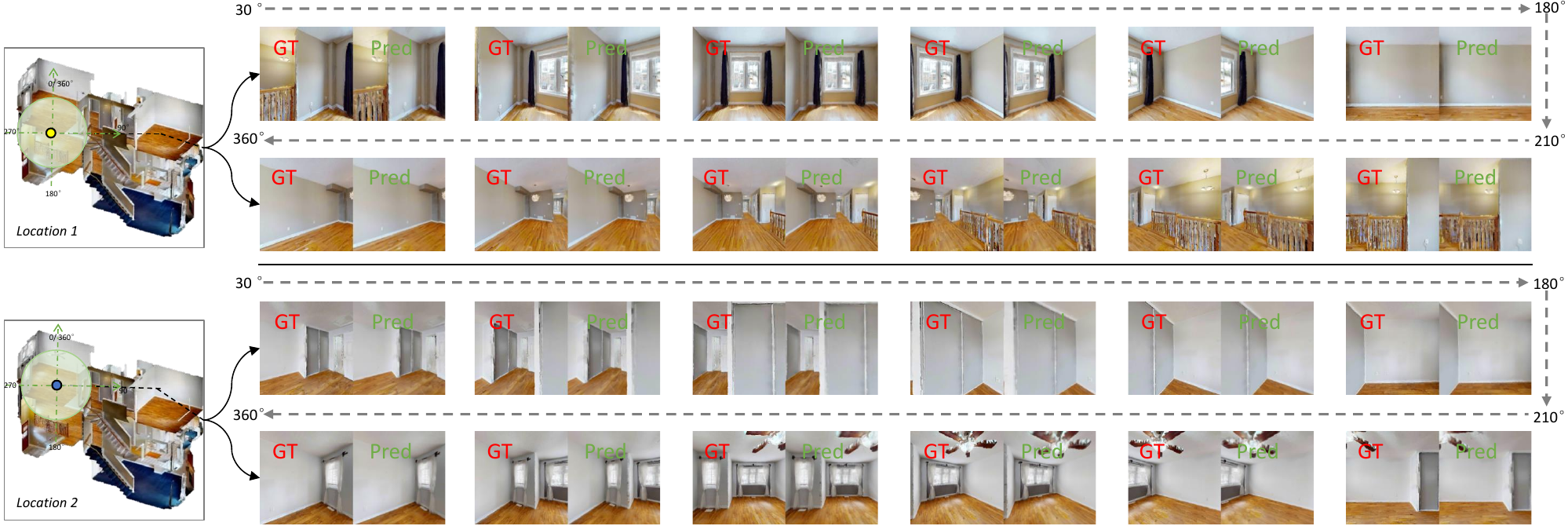}
    \caption{Visualization example of RGB reconstruction for candidate locations using the UnitedVLN model. ``{\color{red}GT}'' and ``{\color{green}Pred}'' denote ground-truth images and rendered images by our pre-training method, respectively.}
    \label{fig:appendix1}
\vspace{-10pt}
\end{figure*}
\vspace{-8pt}

\vspace{-6pt}
\paragraph{Effect on generalizability to other VLN-CE model.}
Table~\ref{table: generalization} illustrates the performance of our proposed 3DGS-based paradigms to generalize two recent-and-representative VLN-CE models, \textit{i.e.}, ETPNav~\cite{an2023etpnav} and HNR~\cite{Wang2024lookahead}, when train the agent to execute VLN-CE task. 
Compared with B1, our B2 (UnitedVLN$_{\text{ETPNav}}$) assembled the future environment with visual images and semantic features achieves significant performance gains for all metrics. Similarly, it also improves the navigation performance when generalizing our 3DGS-based paradigms to HNR model. It proves that our proposed 3DGS-based paradigms can generalize other VLN-CE models and achieve performance gains for them. 
% In addition, our 3DGS-based paradigm can also work on the DTU~\cite{jensen2014dtu} dataset, a large-scale multi-view benchmark for the task of scene reconstruction, when the given point clouds are precise enough. \textit{Please see for more details in supplementary material.}

\begin{table}[t!]
\noindent\begin{minipage}[t]{1\columnwidth}%
\tabcolsep=0.14cm
\begin{center}
\begin{tabular}{ccccc||ccccc}
\Xhline{2.5\arrayrulewidth} % 顶部加粗横线
\#\quad & $\mathcal{L}_{1}^{r}$ & $\mathcal{L}_{2}^{r}$ & $\mathcal{L}_{ssim}^{r}$ & $\mathcal{L}_{2}^{f}$ & 
NE\textdownarrow{} & OSR\textuparrow{} & SR\textuparrow{} & SPL\textuparrow{}\tabularnewline
\hline \hline

E1:\quad &  \ding{51} & \ding{55} & \ding{55} & \ding{55} &  4.68 & 67.9 & 60.4 & 47.9
\tabularnewline

E2:\quad & \ding{55} & \ding{51} & \ding{55} & \ding{55} & 4.59 & 67.7 & 60.5 & 47.6
\tabularnewline

E3:\quad & \ding{51} & \ding{51} & \ding{55} & \ding{55} & 4.47 & 68.2 & 60.7 & 48.1\tabularnewline
 
E4:\quad &  \ding{51} & \ding{51} & \ding{51} & \ding{55} & 4.33 & 68.7 & 61.6 & 50.6\tabularnewline

\hline 
E5:\quad & \ding{51} & \ding{51} & \ding{51} & \ding{51} & \textbf{4.29} & \textbf{70.0} & \textbf{62.4} & \textbf{51.1} \tabularnewline
\hline 
\end{tabular}
\par\end{center}%
\end{minipage}
\vspace{-5pt}
\caption{Ablation study of the different loss in UnitedVLN model.}\label{tab: multi_loss}
\vspace{-10pt}
\end{table}

\begin{figure}[t!]
\vspace{-4pt}
    \centering
    \includegraphics[width=.62\linewidth]{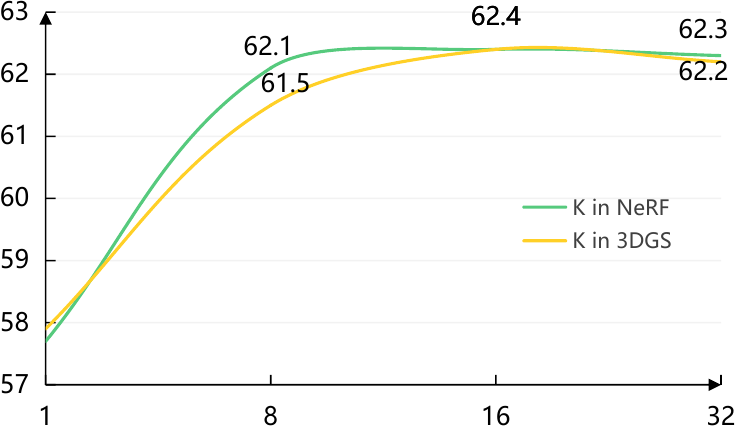}
    \vspace{-8pt}
    \caption{Ablation study of numbers of K in NeRF and 3DGS.}
    \label{fig: K_neast}
\vspace{-10pt}
\end{figure}

\vspace{-8pt}
\paragraph{Effect on speed of image rendering.}
Table~\ref{tab: time} illustrates speed comparisons of image rendering of our UntedVLN with HNR (the SOTA model)  on the stage of pre-training and training. For a fair comparison, we fix the same experiment setting with HNR.  As shown in Table~\ref{tab: time}, UntedVLN rendering speed is about 63$\times$ faster than HNR for all stages: our 0.036s (27.78Hz) \textit{vs.} HNR 2.21s (0.452Hz) for the pre-training stage; our 0.032s (31.25Hz) \textit{vs.} HNR 2.11s (0.473Hz) for the training stage. It proves that UnitedVLN achieves better visual representation with faster rendering. % Here, C1 denotes the HNR model speed of image rendering on pre-training and training.
\vspace{-12pt}
\paragraph{Effect of different feature extractors on rendering.}
Table~\ref{tab: feature extractors} illustrates the performance comparison of UnitedVLN using different feature extractors to encode the rendered image and feature map (\textit{cf.} Eq.~\ref{eq: render-image}) to feature embedding. Here, D1 (ImageNet-ViT-B/16) and D2 (ViT-B/16-CLIP) denote performance using the different pre-trained dataset, \textit{i.e.}, ImageNet~\cite{deng2009imagenet} and CLIP~\cite{radford2021learning}. As shown in Table~\ref{tab: feature extractors}, D2 achieves better performance compared with D1. The reason for performance gain on D2 may CLIP encodes more semantics due to large-scale image-text matching while lacking diverse visual concepts on ImageNet. Thus, we use ViT-B/16-CLIP as the feature extractor, enhancing the semantics of navigation representation.
\vspace{-12pt}
\paragraph{Effect of multi-loss on rendering.}
Table~\ref{table: generalization} illustrates the performance of the proposed UnitedVLN using diverse losses to pre-train, \textit{i.e.}, $\mathcal{L}_{1}^{r}$, $\mathcal{L}_{2}^{r}$, $\mathcal{L}_{ssim}^{r}$, and $\mathcal{L}_{2}^{f}$. Among them, \textit{i.e.}, $\mathcal{L}_{1}^{r}$, and $\mathcal{L}_{2}^{r}$, $\mathcal{L}_{ssim}^{r}$ are adopted use optime rendered RGB images for colors and geometry, while $\mathcal{L}_{2}^{f}$ is for feature similarity. Here, E1 - E5  denote compositions using different loss functions, and E5 achieves the best performance due to jointly optimizing RGB images with better colors and geometric structure, and features with semantics.

\vspace{-5pt}
\subsection{Qualitative Analysis}
\label{Additional results and visualization}
\paragraph{Visualization Example of rendering.} 
\vspace{-2pt}
To validate the effect of pre-training UnitedVLN on rendering image quality, we visualize several 360° panoramic observations surrounding its current location (\textit{i.e.}, 12 view images with 30° separation each). Here, we report each view comparison between rendered images and ground-truth images, as shown in Figure{~\ref{fig:appendix1}}. As shown in Figure{~\ref{fig:appendix1}}, the rendered image not only reconstructs the colors and geometry of the real image but even the bright details of the material (\textit{e.g.}, reflections on a smooth wooden floor). This proves the effect of pre-training UnitedVLN for generalizing high-quality images.
\vspace{-20pt}
\paragraph{Visualization Example of Navigation.} 
\vspace{-3pt}
To validate the effect of UnitedVLN for effective navigation, we report the comparison of navigation strategy between the baseline model (revised ETPNav) and Our UnitedVLN. Here, we also report each node navigation score for a better view. As shown in Figure{~\ref{fig:vis}}, the baseline model suffers navigation error as it obtains limited observations by relying on current waypoint while our UnitedVLN achieves correct decision-marking of navigation by full future explorations that aggregate intuitive appearances and complicated semantics information. This proves the effect of RGB-united-feature future representations, improving the performance of VLN-CE.
\vspace{-6pt}
\section{Conclusion and Discussion}
\label{conclusion}
\vspace{-6pt}
\paragraph{Conclusion.}
We introduce UnitedVLN, a generalizable 3DGS-based pre-training paradigm for improving Continuous Vision-and-Language Navigation (VLN-CE). It pursues full future environment representations by simultaneously rendering the visual images the semantic features with higher-quality 360° from sparse neural points. UnitedVLN has two insightful schemes, \textit{i.e.}, Search-Then-Query sampling (STQ) scheme, and separate-then-united rendering (STU) scheme. For improving model efficiency, STQ searches for each point only in its neighborhood and queries its K-nearest points. For improving model robustness, STU aggregate appearance by splatting in 3DGS and semantics information by volume-rendering in NeRF for robust navigation. Extensive experiments on two VLN-CE benchmarks demonstrate that UnitedVLN significantly outperforms state-of-the-art models.

\vspace{-14pt}
\paragraph{Discussion.}
Some recent work (\textit{e.g.}, HNR~\cite{Wang2024lookahead}) share the same spirit of using future environment rendering but there are some distinct differences: 1) There are different paradigms (\textbf{3DGS} \textit{vs.} \textbf{NeRF}).  2) There are different future explorations ({\textbf{RGB-united-feature} \textit{vs.} \textbf{feature}}), where continuous image prediction in a way that humans can understand makes the agent's behaviors in each step more interpretable. 3) There are different scalability (\textbf{efficient-and-fast rendering} \textit{vs.} \textbf{inefficient-and-slow rendering}). For the proposed UnitedVLN framework, this is a new paradigm that focuses on the full future environment representation besides just rough on single-and-limited features or images. Both STQ and STU are plug-and-play for enforcing sampling and rendering. This well matches our intention, contributing feasible modules like STQ and STU in the embodied AI. One more thing, we hope this paradigm can encourage further investigation of the idea ``\underline{dreaming future} and \underline{doing now}''.

{
\small
\bibliographystyle{ieeenat_fullname}
\bibliography{main}
}

% 补充材料
\clearpage
\setcounter{page}{1}
\maketitlesupplementary
\appendix

\textit{This document provides more details of our method, experimental details and visualization examples, which are organized as follows:}
\begin{itemize}
\item \textit{Model Details (\textit{cf.} §~\ref{Method details})};
\item \textit{Experiment Details (\textit{cf.} §~\ref{Experimental Details})}; 
\item \textit{Visualization Example (\textit{cf.} §~\ref{Additional results and visualization})}.
\item \textit{Discussion Analysis (\textit{cf.} §~\ref{Discussion Analysis})}.
\end{itemize}
{\textit{The \underline{\textbf{anonymous code}} of UnitedVLN is available: \tt\small \url{https://anonymous.4open.science/r/UnitedVLN-08B6/}}}

\section{Model Details}
\label{Method details}
\subsection{Details of Feature Cloud.}
Feature cloud $\mathcal{M}$ focuses on fine-grained contexts of observed environments, consisting of neural-point positions and grid features of features map. Following HNR~\cite{Wang2024lookahead}, we use a pre-trained CLIP-ViT-B~\cite{radford2021learning} model to extract grid features $\{\boldsymbol{h}_{t,i}\in\mathcal{\mathbb{R}}^{H^{'}\times W^{'}\times D^{'}}\}_{i=1}^{12}$, for 12 observed RGB images $\mathcal{R}_{t}=\{r_{t,i}\}_{i=1}^{12}$ at time step $t$. Here, $H^{'}\times W^{'}\times D^{'}$ denotes the resolution of feature maps extracted by CLIP. For the sake of calculability, we omit all the subscripts $(i,h^{'},w^{'})$ and denote its as $u$, where $u$ ranges from 1 to $U$, and $U=12 \cdot$$H^{'}$$\cdot$$W^{'}$.  Then, each grid feature $\boldsymbol{g}_{t,u}\in {\mathbb{R}}^{D}$ in $\mathcal{M}$ is mapped to its 3D world position $\boldsymbol{Q}_{t,j}=[q_x,q_y,q_z]$ following the mapping in Eq.~\ref{maping:projection}. 
Besides the horizontal orientation $\theta_{t,j}$, we also calculate size grid feature scale $s_{t,j}$ using camera's horizontal field-of-view $\Theta_{HFOV}$, as follows
\begin{align}
\boldsymbol{s}_{t,u}=1/W^{'} \cdot [\tan(\Theta_{HFOV}/2)\cdot \boldsymbol{d}_{t,u}],
\end{align}
\noindent where $W^{'}$ is the width of the features maps extracted by the CLIP-ViT-B for each image. In this way, all these grid features and their spatial position of feature maps are perceived in the feature cloud $\mathcal{M}$:
\begin{equation}
    \label{eq:fcd}
    \mathcal{M}_{t}= \mathcal{M}_{t-1} \cup \{[\boldsymbol{Q}_{t,u},\boldsymbol{H}_{t,u}, \boldsymbol{\theta}_{t,u},\boldsymbol{s}_{t,u}]\}_{j=1}^{U}.
\end{equation}

\subsection{Details of Pcd U-Net.}
As shown in Fig.~\ref{fig:appendix1}, we utilize a multi-input and single-output UNet-like architecture (Pcd U-Net) to encode points in the point cloud with different scales to obtain neural descriptors. Specifically, the UNet-like architecture has three set abstractions (SA) and three feature propagations (FP) including multilayer perceptron (MLP), point-voxel convolution (PVC)~\cite{liu2019point}, Grouper block (in SA)~\cite{qi2017pointnet++}, and Nearest-Neighbor-Interpolation (NNI). Through the above modules, we downsample the original point cloud with decreasing rates and concatenate the downsampled point clouds with different levels of feature maps as extra inputs.  Thus, the neural descriptors $\boldsymbol{F}_{p}^{'}$ can formulated as, 
\begin{equation}
\begin{aligned}
    \boldsymbol{F}_{p}^{'} = \mathcal{U}((\boldsymbol{P}^{'},\boldsymbol{C}^{'}), (\boldsymbol{P}^{'},\boldsymbol{C}^{'})_{\downarrow r_1}, (\boldsymbol{P}^{'},\boldsymbol{C}^{'})_{\downarrow r_2}),
    \label{eq: extractor}
\end{aligned}
\end{equation}
where the $\mathcal{U}$ denotes UNet-based extractor network for point cloud. And, the $\boldsymbol{P}^{'}$ and $\boldsymbol{C}^{'}$ denote sampled point coordinates and colors (\textit{cf.} in Eq.~\ref{eq: sample points}). The $\downarrow$ represents a uniform downsampling operation and $r$ denotes the sampling rate where $r_1 > r_2$. This operation allows the extractor to identify features across various scales and receptive fields. The obtained single output feature vector serves as the descriptor for all points.

After that, with the obtained neural descriptors, coordinates as well as colors, we use different heads to regress the corresponding Gaussian properties in a point-wise manner. As shown in Fig.~\ref{fig:appendix1}, the image Gaussian regressor contains three independent heads, such as convolutions and corresponding activation functions, \textit{i.e.}, rotation quaternion ($\boldsymbol{R}$), scale factor ($\boldsymbol{S}$), and opacity ($\boldsymbol{\alpha}$). Meanwhile, in this way, we use the neural descriptors $\boldsymbol{P}^{'}$ to replace colors and obtain feature Gaussians, following general point-rendering practice~\cite{wang2024pfgs}. Thus, the images Gaussians $\mathcal{G_{\scalebox{0.6}{$\triangle$}}}$ and feature Gaussians $\mathcal{G_{\scalebox{0.8}{$\triangledown$}}}$ in the 3DGS branch can be formulated as,
\begin{equation}
    \mathcal{G_{\scalebox{0.6}{$\triangle$}}},\mathcal{G_{\scalebox{0.8}{$\triangledown$}}} = \{\boldsymbol{R}, \boldsymbol{S}, \alpha, \boldsymbol{P}^{'}, \boldsymbol{C}{'}  \},\{\boldsymbol{R}, \boldsymbol{S}, \alpha, \boldsymbol{P}^{'}, \boldsymbol{F}_{p}^{'} \}.
    \label{eq: regressor}
\end{equation}

\subsection{Details of Volume Rendering.}
We set the k-nearest search radius $R$ as 1 meter, and the radius $\hat{R}$ for \textit{sparse sampling} strategy is also set as 1 meter. The rendered ray is uniformly sampled from 0 to 10 meters, and the number of sampled points is set as 256.

\begin{figure*}[t!]
    \centering
    \includegraphics[width=\linewidth]{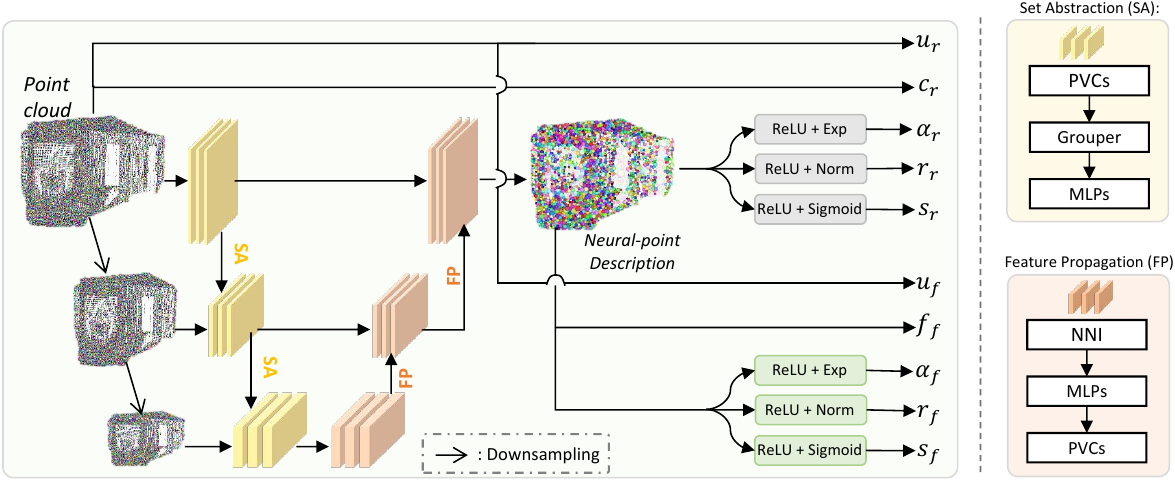}
    \caption{Architecture of Point cloud U-Net. It mainly consists of four stages: point cloud downsampling; set abstractions (SA); point-voxel convolution (PVC) and Gaussian properties prediction.}
    \label{fig:appendix1}
    \vspace{-10pt}
\end{figure*}

\subsection{Details of Gaussian Splatting.}
3DGS is an explicit representation approach for 3D scenes, parameterizing the scene as a set of 3D Gaussian primitives, where each 3D Gaussian is characterized by a full 3D covariance matrix $\Sigma$, its center point $x$, and the spherical harmonic ($SH$). The Gaussian's mean value is expressed as:
\begin{equation}
    G(x)=e^{-\frac{1}{2}(x)^{T}\Sigma^{-1}(x)}.
\end{equation}
To enable optimization via backpropagation, the covariance matrix $\Sigma$ could be decomposed into a rotation matrix ($R$) and a scaling matrix ($S$), as 
\begin{equation}
    \Sigma=RSS^{T}R^{T}.
\end{equation}
Given the camera pose, the projection of the 3D Gaussians to the 2D image plane can be characterized by a view transform matrix ($W$) and the Jacobian of the affine approximation of the projective transformation ($J$), as,
\begin{equation}
    \Sigma^{'}=JW\Sigma W^{T}J^{T}
\end{equation}
where the $\Sigma^{'}$ is the covariance matrix in 2D image planes. Thus, the $\alpha$-blend of  $\mathcal{N}$ ordered points overlapping a pixel is utilized to compute the final color $C$ of the pixel:
\begin{equation}
    C = \sum_{i\in \mathcal{N}}c_i\alpha_i\prod_{j=1}^{i-1}(1-\alpha_j)
    \vspace{-1mm}
\end{equation}
where $c_{i}$ and $\alpha_{i}$ denote the color and density of the pixel with corresponding Gaussian parameters.
\subsection{Objective Function}
According to the stage of VLN-CE, UnitedVLN mainly has two objectives, \textit{i.e.}, one aims to achieve better render quality of images and features (\textit{cf.} Eq~\ref{eq: render-image})  in the pre-training stage and the other is for better navigation performance (\textit{cf.} Eq~\ref{eq: navigation-logits}) in the training stage. The details are as follows.
\paragraph{Pre-training Loss.} For the rendered image (\textit{cf.} Eq.~\ref{eq: render-image}), we use $\mathcal{L}_{1}^{r}$ and $\mathcal{L}_{2}^{r}$ loss to constrain the color similarity between the ground-truth image by calculating its L1 distance and L2 distance in the pixel domain. In addition, we also use SSIM loss $\mathcal{L}_{ssim}^{r}$ to constrain geometric structures between rendered images and ground-truth images. And, the aggregated rendered features (\textit{cf. }Eq.~\ref{nerf-features} and Eq.~\ref{gs-features}) use $\mathcal{L}_{2}^{f}$ to constrain cosine similarity between ground-truth features extracted by the ground-truth image with pre-trained image encoder of CLIP. Thus, all pre-training loss in this paper can be formulated as follows:
\begin{equation}
    \mathcal{L}_{pretrain}=\mathcal{L}_{1}^{r} + \mathcal{L}_{2}^{r} +\mathcal{L}_{ssim}^{r} + \mathcal{L}_{2}^{f}.
    \label{pretrain-loss}
\end{equation}
\paragraph{Training Loss.} For training VLN-CE agent, we use the pre-trained model of the previous stage to generalize future visual images and features in an inference way. Following the practice of previous methods~\cite{Wang2024lookahead,wang2024simtoreal}, with the soft target supervision $\mathcal{A}_{soft}$, the goal scores in navigation (\textit{cf.} in Eq.~\ref{eq: navigation-logits}) are constrained by the cross-entropy (CE) loss:
\vspace{-4pt}
\begin{align}
\mathcal{L}_{nav}=\texttt{CE}(S^{path},\mathcal{A}_{soft}).
\end{align}

% \begin{figure*}[t!]
%     \centering
%     \includegraphics[width=.98\linewidth]{Figs/appendix2.pdf}
%     \caption{Visualization example of RGB reconstruction for candidate locations using the UnitedVLN model. ``{\color{red}GT}'' and ``{\color{green}Pred}'' denote ground-truth images and rendered images by our pre-training method, respectively.}
%     \label{fig:appendix1}
% \vspace{-8pt}
% \end{figure*}

\begin{figure*}[t!]
\vspace{-2pt}
    \centering
    \includegraphics[width=.98\linewidth]{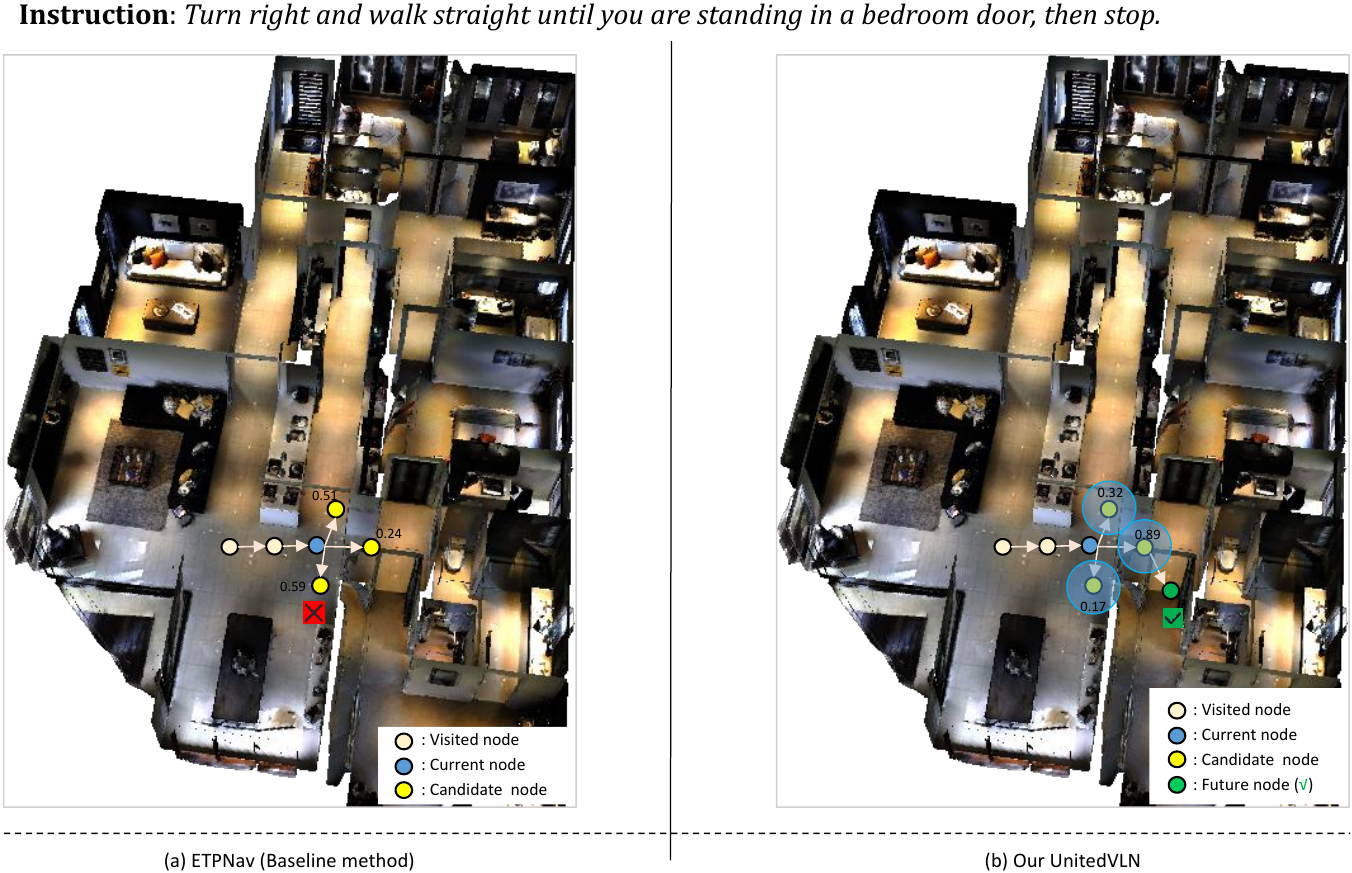}
    \caption{Visualization example of navigation strategy on the val unseen split of the R2R-CE dataset. (a) denotes the navigation strategy of ETPnav (Baseline method). (b) denotes the RGB-united-Feature exploration strategy of our unitedVLN.}
    \label{fig:appendix4}
    \vspace{-8pt}
\end{figure*}

\vspace{-8pt}
\section{Experimental Details}
\label{Experimental Details}
\subsection{Datasets}
To improve the rendered quality of images and features, we first pre-train the proposed 3DGS-based UnitedVLN on the large-scale indoor HM-3D dataset. Following the practice of prior VLN-CE works~\cite{Hong2022bridging,wang2023dreamwalker,an2023etpnav}, 
we evaluate our UnitedVLN two VLN-CE public benchmarks, \textit{i.e,} R2R-CE~\cite{Krantz2020r2r-ce} and RxR-CE~\cite{2020_RXR}.

\textbf{R2R-CE}~\cite{Krantz2020r2r-ce} is derived from the
discrete Matterport3D environments~\cite{matterport3d} based on the Habitat simulator~\cite{ramakrishnan2021habitat}, ensuring the VLN agent to navigate in the continuous environments. The horizontal field-of-view and the turning angle are $90^{\circ}$ and $15^{\circ}$, respectively. It provides step-by-step language instructions, and the average length of instructions is 32 words. 

\textbf{RxR-CE}~\cite{2020_RXR} is an extensive multilingual VLN dataset, containing 126K instructions across English, Hindi, and Telugu. The dataset includes diverse trajectory lengths, averaging 15 meters, making navigation in continuous environments more challenging. In this dataset, The horizontal field-of-view and the turning angle are $79^{\circ}$ and $30^{\circ}$, respectively.

\subsection{Settings of Pre-training.} 
The UnitedVLN model is pre-trained in large-scale HM3D~\cite{ramakrishnan2021habitat} dataset with 800 training scenes. Specifically, we randomly select a starting location in the scene and randomly move to a navigable candidate location at each step. At each step, up to 3 unvisited candidate locations are randomly picked to predict a future view in a random horizontal orientation and render semantic features via NeRF and image/feature via 3DGS. 

\begin{figure*}[t!]
    \centering
    \includegraphics[width=.98\linewidth]{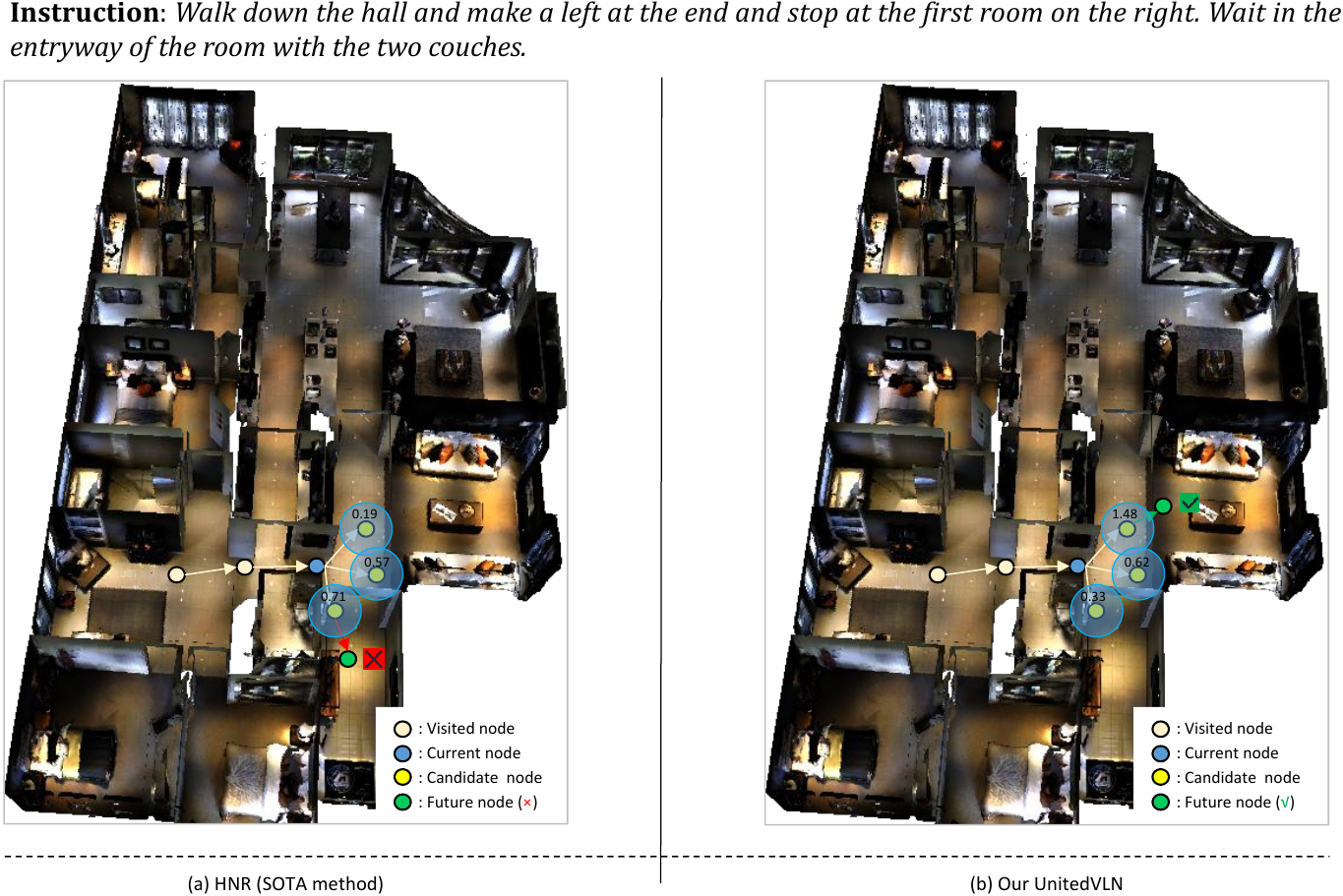}
    \caption{Visualization example of navigation strategy on the val unseen split of the R2R-CE dataset. (a) denotes the navigation strategy of HNR (SOTA method). (b) denotes the RGB-united-Feature exploration strategy of our unitedVLN.}
    \label{fig:appendix3}
    \vspace{-15pt}
\end{figure*}

On the 3DGS branch, the resolution of rendered images and features are $224\times224\times3$ and $224\times224\times768$, which are then fed to the visual encoder of CLIP-ViT-B/16~\cite{radford2021learning} to extract corresponding feature embeddings, \textit{i.e.}, $\boldsymbol{f}_{g}^{r} \in \mathbb{R}^{1\times768}$ and $\boldsymbol{f}_{g}^{f}\in \mathbb{R}^{1\times768}$. Similarly, we use the encoder $\varphi_\text{f}$ of Transformer~\cite{vanswani2017transformer} to extract the feature embedding $\boldsymbol{f}_n\in \mathbb{R}^{1\times768}$ in NeRF. During pre-training, the horizontal field-of-view of each view is set as 90$^{\circ}$. The maximum number of action steps per episode is set to 15. Using 8 Nvidia Tesla A800 GPUs, the UnitedVLN model is pre-trained with a batch size of 4 and a learning rate 1e-4 for 20k episodes.

\subsection{Settings of the Training.}
\paragraph{Settings of R2R-CE dataset.} For R2R-CE, we revise ETPNav~\cite{an2023etpnav} model as our baseline VLN model, where the baseline model initializes with the parameters of ETPNav model trained in the R2R-CE dataset. The UnitedVLN model is trained over 20k episodes on 4 NVIDIA Tesla A800 GPUs, employing a batch size of 8 and a learning rate of 1e-5.
\vspace{-12pt}
\paragraph{Settings of RxR-CE dataset.} Similarly, in RxR-CE, the baseline VLN model is initialized with the parameters of ETPNav~\cite{an2023etpnav} model trained in the RxR-CE dataset. The UnitedVLN model is trained over 100k episodes on 4 NVIDIA Tesla A800 GPUs, employing a batch size of 8 and a learning rate of 1e-5.

\section{Visualization Example}
\label{Additional results and visualization}
% \paragraph{Visualization Example of Pre-training.} 
% To validate the effect of pre-training UnitedVLN on rendering image quality, we visualize several 360° panoramic observations surrounding its current location (\textit{i.e.}, 12 view images with 30° separation each), at the inference stage during pre-training. Here, we also report each view comparison between rendered images and ground-truth images, as shown in Figure{~\ref{fig:appendix1}}. As shown in Figure{~\ref{fig:appendix1}}, the rendered image not only reconstructs the colors and geometry of the real image but even the bright details of the material (\textit{e.g.}, reflections on a smooth wooden floor). This proves the effect of pre-training UnitedVLN for generalizing high-quality images in an inference way.
% \vspace{-10pt}
\paragraph{Visualization Example of Navigation.} 
To validate the effect of UnitedVLN for effective navigation in a continuous environment, we report the visualization comparison of navigation strategy between the baseline model (revised ETPNav) and Our UnitedVLN. Here, we also report each node navigation score for a better view, as shown in Figure{~\ref{fig:appendix4}}. As shown in Figure{~\ref{fig:appendix4}}, the baseline model achieves navigation error as obtains limited observations by relying on a pre-trained waypoint model~\cite{Hong2022bridging} while our UnitedVLN achieves correct decision-marking of navigation by obtaining full future explorations by aggregating intuitive appearances and complicated semantics information. In addition, we also visualize a comparison of navigation strategy between the HNR~\cite{Wang2024lookahead} (SOTA method) and Our UnitedVLN. Compared to HNR relying on limited future features, our UnitedVLN instead aggregates intuitive appearances and complicated semantics information for decision-making and achieving correct navigation. This proves the effect of RGB-united-feature future representations, improving the performance of VLN-CE.

\section{Discussion Analysis}
\label{Discussion Analysis}
\paragraph{Discuss the limitations of 3DGS.}
While 3DGS enables efficient rendering through view-space z-coordinate sorting, this sorting mechanism can introduce popping artifacts (sudden color changes) during camera rotations due to shifts in relative primitive depths. Additionally, Gaussian switching may lead to loss of fine details. In the future, we will explore more advanced Gaussian representations and optimization techniques to avoid popping artifacts and Gaussian switching.
\vspace{-10pt}
\paragraph{Discuss the role of rendering images.}
In our method, it is possible to render semantic features directly using 3DGS or NeRF but achieve suboptimal performance. Our design stems from three key findings: 1) Although the predicted RGB images and feature maps are finally encoded to features, the features after RGB encoder and directly rendered features are similar in semantics but differ in appearance details (RGB retains more complete appearance-level details compared with high-level features of directly using 3DGS/NeRF rendering)  2) Our performance gain mainly drives from multi-source representation fusion (\textit{cf.} Table~\ref{table:each-commpont} of ``Base + STQ + STU"), \textit{i.e.,} low-level RGB (in 3DGS), middle-level semantics (in 3DGS), and high-level semantics (in NeRF), which improves the agent robustness for complicated environments 3) The future visual images created by 3DGS explain the intention of UnitedVLN in a way that human can understand, making agent behaviors more interpretable in each navigation step of decision-making.

\end{document}

%% file: main.bbl
\begin{thebibliography}{54}
\providecommand{\natexlab}[1]{#1}
\providecommand{\url}[1]{\texttt{#1}}
\expandafter\ifx\csname urlstyle\endcsname\relax
  \providecommand{\doi}[1]{doi: #1}\else
  \providecommand{\doi}{doi: \begingroup \urlstyle{rm}\Url}\fi

\bibitem[An et~al.(2022)An, Wang, Li, Wang, Hong, Huang, Wang, and Shao]{an20221st}
Dong An, Zun Wang, Yangguang Li, Yi Wang, Yicong Hong, Yan Huang, Liang Wang, and Jing Shao.
\newblock 1st place solutions for rxr-habitat vision-and-language navigation competition.
\newblock \emph{arXiv preprint arXiv:2206.11610}, 2022.

\bibitem[An et~al.(2023{\natexlab{a}})An, Qi, Li, Huang, Wang, Tan, and Shao]{an2023bevbert}
Dong An, Yuankai Qi, Yangguang Li, Yan Huang, Liang Wang, Tieniu Tan, and Jing Shao.
\newblock Bevbert: Multimodal map pre-training for language-guided navigation.
\newblock In \emph{ICCV}, pages 2737--2748, 2023{\natexlab{a}}.

\bibitem[An et~al.(2023{\natexlab{b}})An, Wang, Wang, Wang, Huang, He, and Wang]{an2023etpnav}
Dong An, Hanqing Wang, Wenguan Wang, Zun Wang, Yan Huang, Keji He, and Liang Wang.
\newblock Etpnav: Evolving topological planning for vision-language navigation in continuous environments.
\newblock \emph{arXiv preprint arXiv:2304.03047}, 2023{\natexlab{b}}.

\bibitem[Anderson et~al.(2018)Anderson, Wu, Teney, Bruce, Johnson, S{\"u}nderhauf, Reid, Gould, and Van Den~Hengel]{VLN_2018vision}
Peter Anderson, Qi Wu, Damien Teney, Jake Bruce, Mark Johnson, Niko S{\"u}nderhauf, Ian Reid, Stephen Gould, and Anton Van Den~Hengel.
\newblock Vision-and-language navigation: Interpreting visually-grounded navigation instructions in real environments.
\newblock In \emph{CVPR}, pages 3674--3683, 2018.

\bibitem[Chang et~al.(2017)Chang, Dai, Funkhouser, Halber, Niessner, Savva, Song, Zeng, and Zhang]{matterport3d}
Angel Chang, Angela Dai, Thomas Funkhouser, Maciej Halber, Matthias Niessner, Manolis Savva, Shuran Song, Andy Zeng, and Yinda Zhang.
\newblock Matterport3d: Learning from rgb-d data in indoor environments.
\newblock In \emph{3DV}, pages 667--676, 2017.

\bibitem[Chen et~al.(2022{\natexlab{a}})Chen, Ji, Lin, Zeng, Li, Tan, and Gan]{chen2022weakly}
Peihao Chen, Dongyu Ji, Kunyang Lin, Runhao Zeng, Thomas~H Li, Mingkui Tan, and Chuang Gan.
\newblock Weakly-supervised multi-granularity map learning for vision-and-language navigation.
\newblock In \emph{Advances in Neural Information Processing Systems}, pages 38149--38161, 2022{\natexlab{a}}.

\bibitem[Chen et~al.(2021)Chen, Guhur, Schmid, and Laptev]{chen2021history}
Shizhe Chen, Pierre-Louis Guhur, Cordelia Schmid, and Ivan Laptev.
\newblock History aware multimodal transformer for vision-and-language navigation.
\newblock In \emph{Advances in Neural Information Processing Systems}, pages 5834--5847, 2021.

\bibitem[Chen et~al.(2022{\natexlab{b}})Chen, Guhur, Tapaswi, Schmid, and Laptev]{chen2022think-GL}
Shizhe Chen, Pierre-Louis Guhur, Makarand Tapaswi, Cordelia Schmid, and Ivan Laptev.
\newblock Think global, act local: Dual-scale graph transformer for vision-and-language navigation.
\newblock In \emph{CVPR}, pages 16537--16547, 2022{\natexlab{b}}.

\bibitem[Deng et~al.(2009)Deng, Dong, Socher, Li, Li, and Fei-Fei]{deng2009imagenet}
Jia Deng, Wei Dong, Richard Socher, Li-Jia Li, Kai Li, and Li Fei-Fei.
\newblock Imagenet: A large-scale hierarchical image database.
\newblock In \emph{CVPR}, pages 248--255, 2009.

\bibitem[DeVries et~al.(2021)DeVries, Bautista, Srivastava, Taylor, and Susskind]{devries2021unconstrained}
Terrance DeVries, Miguel~Angel Bautista, Nitish Srivastava, Graham~W Taylor, and Joshua~M Susskind.
\newblock Unconstrained scene generation with locally conditioned radiance fields.
\newblock In \emph{ICCV}, pages 14304--14313, 2021.

\bibitem[Feng et~al.(2022)Feng, Fu, Lu, and Wang]{feng2022uln}
Weixi Feng, Tsu-Jui Fu, Yujie Lu, and William~Yang Wang.
\newblock Uln: Towards underspecified vision-and-language navigation.
\newblock \emph{arXiv preprint arXiv:2210.10020}, 2022.

\bibitem[Forrester(1971)]{forrester1971counterintuitive}
Jay~W Forrester.
\newblock Counterintuitive behavior of social systems.
\newblock \emph{Theory and Decision}, 2\penalty0 (2):\penalty0 109--140, 1971.

\bibitem[Fried et~al.(2018)Fried, Hu, Cirik, Rohrbach, Andreas, Morency, Berg-Kirkpatrick, Saenko, Klein, and Darrell]{2018-speaker}
Daniel Fried, Ronghang Hu, Volkan Cirik, Anna Rohrbach, Jacob Andreas, Louis-Philippe Morency, Taylor Berg-Kirkpatrick, Kate Saenko, Dan Klein, and Trevor Darrell.
\newblock Speaker-follower models for vision-and-language navigation.
\newblock In \emph{Advances in Neural Information Processing Systems}, 2018.

\bibitem[Georgakis et~al.(2022)Georgakis, Schmeckpeper, Wanchoo, Dan, Miltsakaki, Roth, and Daniilidis]{georgakis2022cm2}
Georgios Georgakis, Karl Schmeckpeper, Karan Wanchoo, Soham Dan, Eleni Miltsakaki, Dan Roth, and Kostas Daniilidis.
\newblock Cross-modal map learning for vision and language navigation.
\newblock In \emph{CVPR}, 2022.

\bibitem[Grandits et~al.(2021)Grandits, Effland, Pock, Krause, Plank, and Pezzuto]{grandits_geasi_2021}
Thomas Grandits, Alexander Effland, Thomas Pock, Rolf Krause, Gernot Plank, and Simone Pezzuto.
\newblock {GEASI}: {Geodesic}-based earliest activation sites identification in cardiac models.
\newblock \emph{International Journal for Numerical Methods in Biomedical Engineering.}, 37\penalty0 (8):\penalty0 e3505, 2021.

\bibitem[Hong et~al.(2021)Hong, Wu, Qi, Rodriguez-Opazo, and Gould]{hong2021vln-bert}
Yicong Hong, Qi Wu, Yuankai Qi, Cristian Rodriguez-Opazo, and Stephen Gould.
\newblock Vln bert: A recurrent vision-and-language bert for navigation.
\newblock In \emph{CVPR}, pages 1643--1653, 2021.

\bibitem[Hong et~al.(2022)Hong, Wang, Wu, and Gould]{Hong2022bridging}
Yicong Hong, Zun Wang, Qi Wu, and Stephen Gould.
\newblock Bridging the gap between learning in discrete and continuous environments for vision-and-language navigation.
\newblock In \emph{CVPR}, 2022.

\bibitem[Hong et~al.(2023)Hong, Zhou, Zhang, Dernoncourt, Bui, Gould, and Tan]{hong2023learning}
Yicong Hong, Yang Zhou, Ruiyi Zhang, Franck Dernoncourt, Trung Bui, Stephen Gould, and Hao Tan.
\newblock Learning navigational visual representations with semantic map supervision.
\newblock In \emph{ICCV}, pages 3055--3067, 2023.

\bibitem[Huang et~al.(2023)Huang, Mees, Zeng, and Burgard]{huang23vlmaps}
Chenguang Huang, Oier Mees, Andy Zeng, and Wolfram Burgard.
\newblock Visual language maps for robot navigation.
\newblock In \emph{ICRA}, London, UK, 2023.

\bibitem[Isola et~al.(2017)Isola, Zhu, Zhou, and Efros]{Isola2021gan}
Phillip Isola, Jun-Yan Zhu, Tinghui Zhou, and Alexei~A. Efros.
\newblock Image-to-image translation with conditional adversarial networks.
\newblock In \emph{CVPR}, pages 5967--5976, 2017.

\bibitem[Johnson-Laird(1983)]{johnson1983mental}
Philip~Nicholas Johnson-Laird.
\newblock \emph{Mental models: Towards a cognitive science of language, inference, and consciousness}.
\newblock Harvard University Press, 1983.

\bibitem[Johnson-Laird(2010)]{johnson2010mental}
Philip~N Johnson-Laird.
\newblock Mental models and human reasoning.
\newblock \emph{Proceedings of the National Academy of Sciences}, 2010.

\bibitem[Kerbl et~al.(2023)Kerbl, Kopanas, Leimk{\"u}hler, and Drettakis]{kerbl20233d}
Bernhard Kerbl, Georgios Kopanas, Thomas Leimk{\"u}hler, and George Drettakis.
\newblock 3d gaussian splatting for real-time radiance field rendering.
\newblock \emph{TOG}, 42\penalty0 (4):\penalty0 1--14, 2023.

\bibitem[Koh et~al.(2021)Koh, Lee, Yang, Baldridge, and Anderson]{koh2021pathdreamer}
Jing~Yu Koh, Honglak Lee, Yinfei Yang, Jason Baldridge, and Peter Anderson.
\newblock Pathdreamer: A world model for indoor navigation.
\newblock In \emph{ICCV}, pages 14738--14748, 2021.

\bibitem[Krantz and Lee(2022)]{krantz2022sim2sim}
Jacob Krantz and Stefan Lee.
\newblock Sim-2-sim transfer for vision-and-language navigation in continuous environments.
\newblock In \emph{ECCV}, 2022.

\bibitem[Krantz et~al.(2020)Krantz, Wijmans, Majumdar, Batra, and Lee]{Krantz2020r2r-ce}
Jacob Krantz, Erik Wijmans, Arjun Majumdar, Dhruv Batra, and Stefan Lee.
\newblock Beyond the nav-graph: Vision-and-language navigation in continuous environments.
\newblock In \emph{ECCV}, 2020.

\bibitem[Ku et~al.(2020)Ku, Anderson, Patel, Ie, and Baldridge]{2020_RXR}
Alexander Ku, Peter Anderson, Roma Patel, Eugene Ie, and Jason Baldridge.
\newblock Room-across-room: Multilingual vision-and-language navigation with dense spatiotemporal grounding.
\newblock In \emph{EMNLP}, pages 4392--4412, 2020.

\bibitem[Kwon et~al.(2023)Kwon, Park, and Oh]{kwon2023renderable}
Obin Kwon, Jeongho Park, and Songhwai Oh.
\newblock Renderable neural radiance map for visual navigation.
\newblock In \emph{CVPR}, pages 9099--9108, 2023.

\bibitem[Li and Bansal(2023{\natexlab{a}})]{jialu2023vlnsig}
Jialu Li and Mohit Bansal.
\newblock Improving vision-and-language navigation by generating future-view image semantics.
\newblock In \emph{CVPR}, pages 10803--10812, 2023{\natexlab{a}}.

\bibitem[Li and Bansal(2023{\natexlab{b}})]{li2023improving}
Jialu Li and Mohit Bansal.
\newblock Improving vision-and-language navigation by generating future-view image semantics.
\newblock In \emph{CVPR}, pages 10803--10812, 2023{\natexlab{b}}.

\bibitem[Liu et~al.(2024{\natexlab{a}})Liu, Wang, and Yang]{liurui2024energy}
Rui Liu, Wenguan Wang, and Yi Yang.
\newblock Vision-language navigation with energy-based policy.
\newblock \emph{Advances in Neural Information Processing Systems}, 37:\penalty0 108208--108230, 2024{\natexlab{a}}.

\bibitem[Liu et~al.(2024{\natexlab{b}})Liu, Wang, Hu, Shen, Ye, Zang, Cao, Li, and Liu]{liu2024mvsgauss}
Tianqi Liu, Guangcong Wang, Shoukang Hu, Liao Shen, Xinyi Ye, Yuhang Zang, Zhiguo Cao, Wei Li, and Ziwei Liu.
\newblock Mvsgaussian: Fast generalizable gaussian splatting reconstruction from multi-view stereo.
\newblock In \emph{ECCV}, 2024{\natexlab{b}}.

\bibitem[Liu et~al.(2019)Liu, Tang, Lin, and Han]{liu2019point}
Zhijian Liu, Haotian Tang, Yujun Lin, and Song Han.
\newblock Point-voxel cnn for efficient 3d deep learning.
\newblock \emph{Advances in Neural Information Processing Systems}, 32, 2019.

\bibitem[Mildenhall et~al.(2021)Mildenhall, Srinivasan, Tancik, Barron, Ramamoorthi, and Ng]{mildenhall2021nerf}
Ben Mildenhall, Pratul~P Srinivasan, Matthew Tancik, Jonathan~T Barron, Ravi Ramamoorthi, and Ren Ng.
\newblock Nerf: Representing scenes as neural radiance fields for view synthesis.
\newblock \emph{CACM}, 65:\penalty0 99--106, 2021.

\bibitem[Park et~al.(2021)Park, Sinha, Barron, Bouaziz, Goldman, Seitz, and Martin-Brualla]{park2021nerfies}
Keunhong Park, Utkarsh Sinha, Jonathan~T Barron, Sofien Bouaziz, Dan~B Goldman, Steven~M Seitz, and Ricardo Martin-Brualla.
\newblock Nerfies: Deformable neural radiance fields.
\newblock In \emph{CVPR}, pages 5865--5874, 2021.

\bibitem[Pashevich et~al.(2021)Pashevich, Schmid, and Sun]{pashevich2021episodic}
Alexander Pashevich, Cordelia Schmid, and Chen Sun.
\newblock Episodic transformer for vision-and-language navigation.
\newblock In \emph{ICCV}, 2021.

\bibitem[Qi et~al.(2017)Qi, Yi, Su, and Guibas]{qi2017pointnet++}
Charles~Ruizhongtai Qi, Li Yi, Hao Su, and Leonidas~J Guibas.
\newblock Pointnet++: Deep hierarchical feature learning on point sets in a metric space.
\newblock \emph{Advances in neural information processing systems}, 30, 2017.

\bibitem[Qi et~al.(2020)Qi, Wu, Anderson, Wang, Wang, Shen, and Hengel]{2020reverie}
Yuankai Qi, Qi Wu, Peter Anderson, Xin Wang, William~Yang Wang, Chunhua Shen, and Anton van~den Hengel.
\newblock Reverie: Remote embodied visual referring expression in real indoor environments.
\newblock In \emph{CVPR}, pages 9982--9991, 2020.

\bibitem[Radford et~al.(2021)Radford, Kim, Hallacy, Ramesh, Goh, Agarwal, Sastry, Askell, Mishkin, Clark, et~al.]{radford2021learning}
Alec Radford, Jong~Wook Kim, Chris Hallacy, Aditya Ramesh, Gabriel Goh, Sandhini Agarwal, Girish Sastry, Amanda Askell, Pamela Mishkin, Jack Clark, et~al.
\newblock Learning transferable visual models from natural language supervision.
\newblock In \emph{ICML}, pages 8748--8763, 2021.

\bibitem[Ramakrishnan et~al.(2021)Ramakrishnan, Gokaslan, Wijmans, Maksymets, Clegg, Turner, Undersander, Galuba, Westbury, Chang, et~al.]{ramakrishnan2021habitat}
Santhosh~K Ramakrishnan, Aaron Gokaslan, Erik Wijmans, Oleksandr Maksymets, Alex Clegg, John Turner, Eric Undersander, Wojciech Galuba, Andrew Westbury, Angel~X Chang, et~al.
\newblock Habitat-matterport 3d dataset (hm3d): 1000 large-scale 3d environments for embodied ai.
\newblock \emph{arXiv preprint arXiv:2109.08238}, 2021.

\bibitem[Ramesh et~al.(2021)Ramesh, Pavlov, Goh, Gray, Voss, Radford, Chen, and Sutskever]{ramesh2021zero}
Aditya Ramesh, Mikhail Pavlov, Gabriel Goh, Scott Gray, Chelsea Voss, Alec Radford, Mark Chen, and Ilya Sutskever.
\newblock Zero-shot text-to-image generation.
\newblock In \emph{CoRL}, pages 8821--8831, 2021.

\bibitem[Tan et~al.(2019)Tan, Yu, and Bansal]{tan2019learning}
Hao Tan, Licheng Yu, and Mohit Bansal.
\newblock Learning to navigate unseen environments: Back translation with environmental dropout.
\newblock In \emph{NAACL}, pages 2610--2621, 2019.

\bibitem[Thomason et~al.(2020)Thomason, Murray, Cakmak, and Zettlemoyer]{thomason2020cvdn}
Jesse Thomason, Michael Murray, Maya Cakmak, and Luke Zettlemoyer.
\newblock Vision-and-dialog navigation.
\newblock In \emph{PMLR}, 2020.

\bibitem[Vaswani et~al.(2017)Vaswani, Shazeer, Parmar, Uszkoreit, Jones, Gomez, Kaiser, and Polosukhin]{vanswani2017transformer}
Ashish Vaswani, Noam Shazeer, Niki Parmar, Jakob Uszkoreit, Llion Jones, Aidan~N. Gomez, Lukasz Kaiser, and Illia Polosukhin.
\newblock Attention is all you need.
\newblock \emph{Advances in Neural Information Processing Systems}, 33:\penalty0 5998--6008, 2017.

\bibitem[Wang et~al.(2020)Wang, Wang, Shu, Liang, and Shen]{wang2020active}
Hanqing Wang, Wenguan Wang, Tianmin Shu, Wei Liang, and Jianbing Shen.
\newblock Active visual information gathering for vision-language navigation.
\newblock In \emph{ECCV}, pages 307--322, 2020.

\bibitem[Wang et~al.(2023{\natexlab{a}})Wang, Liang, Van~Gool, and Wang]{wang2023dreamwalker}
Hanqing Wang, Wei Liang, Luc Van~Gool, and Wenguan Wang.
\newblock Dreamwalker: Mental planning for continuous vision-language navigation.
\newblock In \emph{ICCV}, pages 10873--10883, 2023{\natexlab{a}}.

\bibitem[Wang et~al.(2024{\natexlab{a}})Wang, Zhang, He, and Xu]{wang2024pfgs}
Jiaxu Wang, Ziyi Zhang, Junhao He, and Renjing Xu.
\newblock Pfgs: High fidelity point cloud rendering via feature splatting.
\newblock \emph{arXiv preprint arXiv:2407.03857}, 2024{\natexlab{a}}.

\bibitem[Wang et~al.(2023{\natexlab{b}})Wang, Wu, Yao, and Wang]{Ting2023graph-vlnce}
Ting Wang, Zongkai Wu, Feiyu Yao, and Donglin Wang.
\newblock Graph based environment representation for vision-and-language navigation in continuous environments.
\newblock \emph{arXiv preprint arXiv:2301.04352}, 2023{\natexlab{b}}.

\bibitem[Wang et~al.(2019)Wang, Huang, Celikyilmaz, Gao, Shen, Wang, Wang, and Zhang]{2019reinforced}
Xin Wang, Qiuyuan Huang, Asli Celikyilmaz, Jianfeng Gao, Dinghan Shen, Yuan-Fang Wang, William~Yang Wang, and Lei Zhang.
\newblock Reinforced cross-modal matching and self-supervised imitation learning for vision-language navigation.
\newblock In \emph{CVPR}, pages 6629--6638, 2019.

\bibitem[Wang et~al.(2023{\natexlab{c}})Wang, Li, Yang, Liu, and Jiang]{wang2023gridmm}
Zihan Wang, Xiangyang Li, Jiahao Yang, Yeqi Liu, and Shuqiang Jiang.
\newblock Gridmm: Grid memory map for vision-and-language navigation.
\newblock In \emph{ICCV}, pages 15625--15636, 2023{\natexlab{c}}.

\bibitem[Wang et~al.(2024{\natexlab{b}})Wang, Li, Yang, Liu, Hu, Jiang, and Jiang]{Wang2024lookahead}
Zihan Wang, Xiangyang Li, Jiahao Yang, Yeqi Liu, Junjie Hu, Ming Jiang, and Shuqiang Jiang.
\newblock Lookahead exploration with neural radiance representation for continuous vision-language navigation.
\newblock In \emph{CVPR}, pages 13753--13762, 2024{\natexlab{b}}.

\bibitem[Wang et~al.(2024{\natexlab{c}})Wang, Li, Yang, Liu, and Jiang]{wang2024simtoreal}
Zihan Wang, Xiangyang Li, Jiahao Yang, Yeqi Liu, and Shuqiang Jiang.
\newblock Sim-to-real transfer via 3d feature fields for vision-and-language navigation.
\newblock In \emph{Conference on Robot Learning (CoRL)}, 2024{\natexlab{c}}.

\bibitem[Xia and Xue(2023)]{3d2022survey}
Weihao Xia and Jing-Hao Xue.
\newblock A survey on deep generative 3d-aware image synthesis.
\newblock In \emph{ACM Comput. Surv.}, pages 1--34, 2023.

\bibitem[Zhu et~al.(2021)Zhu, Liang, Zhu, Yu, Chang, and Liang]{zhu2021soon}
Fengda Zhu, Xiwen Liang, Yi Zhu, Qizhi Yu, Xiaojun Chang, and Xiaodan Liang.
\newblock Soon: Scenario oriented object navigation with graph-based exploration.
\newblock In \emph{CVPR}, pages 12689--12699, 2021.

\end{thebibliography}
